\pdfoutput=1

\documentclass[11pt]{article}

\usepackage[final]{acl}

\usepackage{times}
\usepackage{latexsym}
\usepackage{amsmath}
\usepackage{amssymb}
\usepackage{bbm}
\usepackage{multirow}
\usepackage{caption}
\usepackage{adjustbox}
\usepackage{booktabs}

\usepackage[T1]{fontenc}

\usepackage[utf8]{inputenc}

\usepackage{microtype}
\usepackage{subcaption}
\usepackage{inconsolata}

\usepackage{graphicx}
\usepackage{comment}
\title{Local Contrastive Editing of Gender Stereotypes}

\author{Marlene Lutz \\
  University of Mannheim \\
  \texttt{marlene.lutz@uni-mannheim.de} \\ \And
  Rochelle Choenni \\
  University of Amsterdam \\
  \texttt{r.m.v.k.choenni@uva.nl} \\ \AND
  Markus Strohmaier \\
  University of Mannheim, GESIS, CSH Vienna \\
  \texttt{markus.strohmaier@uni-mannheim.de} \\ \And
  Anne Lauscher \\
  University of Hamburg \\
  \texttt{anne.lauscher@uni-hamburg.de}}

\begin{document}
\maketitle
\begin{abstract}
Stereotypical bias encoded in language models (LMs) poses a threat to safe language techno-logy, yet our understanding of how bias manifests in the parameters of LMs remains incomplete. We introduce \emph{local contrastive editing} that enables the localization and editing of a subset of weights in a target model \emph{in relation} to a reference model. We deploy this approach to identify and modify subsets of weights that are associated with gender stereotypes in LMs.
Through a series of experiments, we demonstrate that local contrastive editing can precisely localize and control a small subset (\textless $0.5\%$) of weights that encode gender bias. 
Our work (i) advances our understanding of how stereotypical biases can manifest in the parameter space of LMs and (ii) opens up new avenues for developing parameter-efficient strategies for controlling model properties in a contrastive manner.
\end{abstract}

\section{Introduction}
Stereotypical bias encoded in language models (LMs) can adversely affect the fairness and inclusivity of language technology applications for all users~\citep{blodgett-etal-2020-language, choenni2021stepmothers, ma-etal-2023-deciphering}. 
While considerable efforts have been devoted to measuring~\citep{nadeem2020stereoset, doi:10.1126/science.aal4230} and mitigating~\citep{lauscher-etal-2021-sustainable-modular} such biases, our understanding of where they manifest in the parameter space of LMs remains limited.
Precisely pinpointing biases within the parameters of LMs could enable the development of more targeted and informed bias mitigation strategies.
While current research \citep{ma-etal-2023-deciphering, meissner-etal-2022-debiasing, hauzenberger-etal-2023-modular} has explored identifying and modifying subcomponents of LMs for bias mitigation, we still lack a thorough understanding of the precise manifestation of biases such as stereotypes in specific model weights. 

 
\paragraph{Research Goal}
Consequently, this work aims to (i) localize individual weights that drive stereotypical gender bias in LMs and (ii) modify these weights to steer and mitigate the bias.

\begin{figure}[!t]
    \centering
    \includegraphics[width=\linewidth]{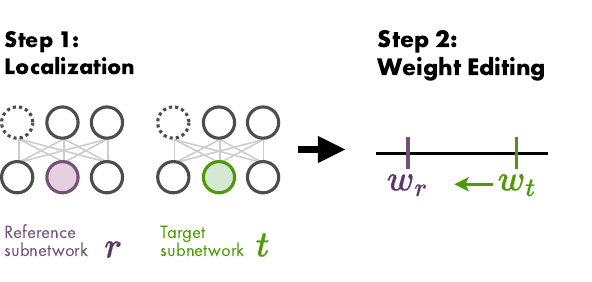}
    \caption{\textbf{Local contrastive editing.} In step 1, we localize weights within a target model that encode a certain property. In Step 2, we modify these selected weights relative to a reference model.}
    \label{fig:main_fig}
\end{figure}

\paragraph{Approach}
We present \emph{local contrastive editing}, a two-step approach that enables the localization and modification of a subset of weights within a target model, relative to a reference model, to control bias (see Figure \ref{fig:main_fig}).
In step 1, we pinpoint individual weights that encode gender stereotypes via unstructured pruning~\citep{chen2020lottery}.
In step 2, we deploy various \emph{local} editing strategies such as weight interpolation or pruning to adjust the identified weights in relation to a reference model.

\paragraph{Results and Contributions}
We demonstrate the feasibility of local constrastive editing for controlling stereotypical gender bias through a series of experiments. Using our approach, we are able to identify subsets of weights that drive stereotypical bias in LMs. We find that our local editing strategies can flexibly steer gender bias while at the same time retaining the functionality of the model. We provide experimental evidence that most strategies enable a smooth and controllable transition of bias between networks and empirically find that a small subset of weights (\textless $0.5\%$) is already sufficient to successfully modify and, finally, mitigate the measurable bias.

\section{Related Work}

\paragraph{Gender Bias} Gender bias is present throughout the entire NLP pipeline, from training data~\citep{leavy2020mitigating} to model representations~\citep{bolukbasi2016man, gonen2019lipstick}, and predictions~\citep{dong2024disclosure}. Consequently, much effort has been put towards locating~\citep{joniak-aizawa-2022-gender, chintam-etal-2023-identifying} and mitigating gender bias at various stages~\citep{sun2019mitigating, lauscher-etal-2021-sustainable-modular, hauzenberger-etal-2023-modular}. 
We study how gender bias manifests in LMs by detecting and editing a minimal set of relevant parameters to control it.

\paragraph{Knowledge Localization} 
Pruning methods have been used to uncover subnetworks, i.e. subsets of model parameters~\citep{frankle2018lottery} isolating \emph{task-specific}~\citep{nooralahzadeh2023improving}, \emph{domain-specific}~\citep{hendy2022domain} or \emph{language-specific}~\citep{wang2020negative, choenni-etal-2023-cross, choenni2023examining, nooralahzadeh2023improving} information. In this paper, we use pruning to find subnetworks that contain stereotypical gender bias. 
Previous research~\citep{vig2020investigating, chintam-etal-2023-identifying} suggests that stereotypical gender bias is concentrated in specific substructures of a network such as attention heads~\citep{chintam-etal-2023-identifying, ma-etal-2023-deciphering, vig2020investigating} or neurons~\citep{vig2020investigating}. We aim to pinpoint the individual weights responsible for encoding gender bias within a network via unstructured pruning~\citep{chen2020lottery}.

\paragraph{Model Editing}

Pretrained LMs serve as backbone for many downstream applications, requiring them to be tailored to specific needs. However, the growing size of language models has made traditional fine-tuning costly, leading to increased interest in alternative refinement methods that avoid gradient updates~\citep{yao-etal-2023-editing}.
One such line of research focuses on efficient model weight editing strategies~\citep{ilharco2022editing, ilharco2022patching, gueta-etal-2023-knowledge}, 
using mathematical operations on weight vectors composed from the full model to modify information. In this paper, we take a more fine-grained approach to model editing, and instead focus on editing only a subset of weights that we identify as being of relevance for encoding gender stereotypical biases beforehand. 

\section{Local Contrastive Editing}
We localize and adjust specific weights in a target model that are responsible for encoding properties such as stereotypical bias. 
To achieve this, we use several contrastive strategies based on comparing a \emph{target network} with a \emph{reference network} that differ in a property of interest.
We refer to this group of techniques as \emph{contrastive weight editing}.

Formally, let $f(x, \theta)$ be the output of a network with parameters $\theta \in \mathbb{R}^d$ for an example input $x$. Given a target network $f_t(\cdot, \theta_t)$ and a reference network $f_r(\cdot, \theta_r)$ of the same architecture, with $\theta_t, \theta_r \in \mathbb{R}^d$, we aim to edit $\theta_t$ w.r.t. $\theta_r$ to modify a property of interest $p$ in $f_t$ while maintaining performance on the original fine-tuning task.

\begin{figure*}[!t]
    \centering
    \includegraphics[width=0.9\linewidth]{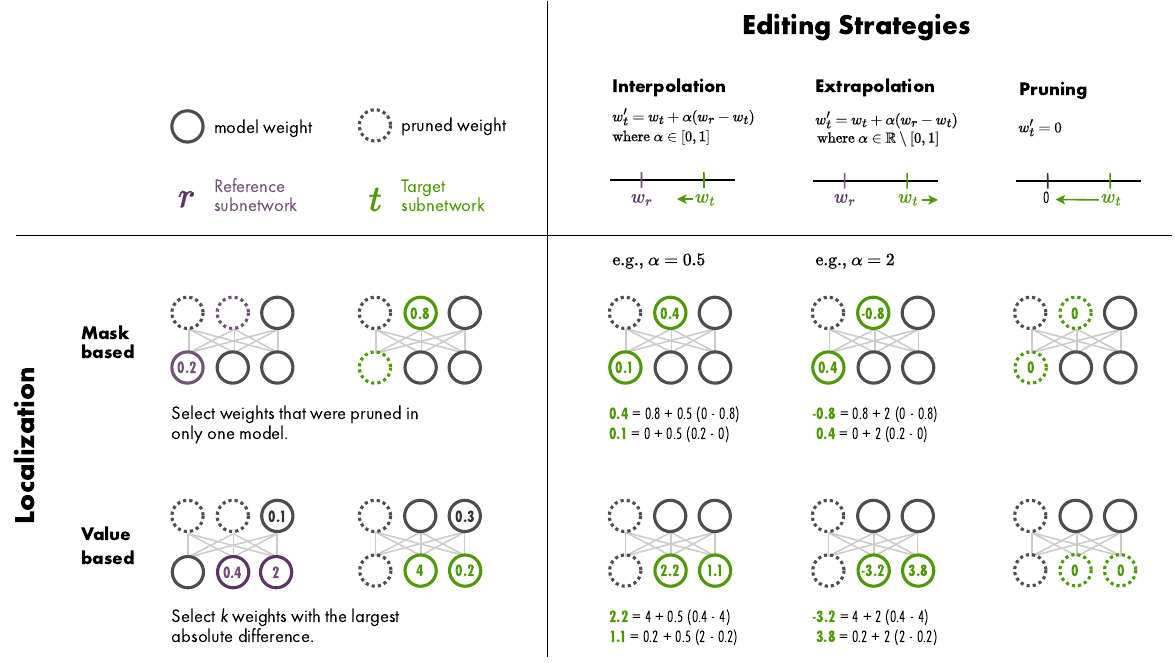}
    \caption{\textbf{Overview of localization and editing strategies.} We show value-based and masked-based localization together with our editing strategies (inter- and  extrapolation, pruning) on exemplary target and reference networks.}
    \label{fig:pipeline}
\end{figure*}

\subsection{Localization} \label{sec:localization}


In the first step, we investigate how a specific property $p$ manifests in the parameter space of a model and try to localize the individual weights associated with it. To this end, we use unstructured magnitude pruning~\cite{chen2020lottery} and discover subnetworks in a target and a reference network that are linked to the encoding of $p$.
We define a subnetwork for a network ${f(\cdot, \theta)}$ as ${f(\cdot, m \odot \theta)}$, where  $\odot$ is the element-wise product and ${m \in \{0, 1\}^d}$ is a binary pruning mask that sets some parameters in $\theta$ to $0$. 
By comparing the target and reference subnetworks, we aim to identify subsets of weights that are related to the encoding of $p$. 
We note that subnetworks extracted from different parent models can differ in two aspects: 
(1) their \emph{pruning masks} may set different parameters to $0$; and 
(2) their parameters $\theta$ may have different \emph{values}.

We explore both aspects separately by first selecting the corresponding subsets of weights and then using them to modify the target network.
To this end, we define a \emph{localization mask} as the outcome of a particular localization strategy, which indicates which weights will be edited in the subsequent step. Formally, we define such a mask for a given index set $\mathcal{I} \subseteq \{1, \dots, d\}$ as
$b := b(\mathcal{I}) \in \{0, 1\}^d$ 
via its elements, such that $b_i = \mathbbm{1} \{i \in \mathcal{I}\}$.
A value of $1$ at index $i$ indicates that the corresponding weight is selected for editing.
We propose the following two strategies to compute such localization masks.

\paragraph{Mask-based Localization}
Given a target subnetwork $f_t(\cdot,  m_t \odot \theta_t)$ and a reference subnetwork $f_r(\cdot,  m_r \odot \theta_r)$, we select those weights that are present in only one of the subnetworks, indicated by the pruning masks $m_t$ and $m_r$ . Formally, we compute the localization mask $b$ as:

\begin{small}
\begin{equation}
    b = m_t \odot  m_r\,.
\end{equation}
\end{small}

\noindent We hypothesize that precisely because weights are pruned in one network, but not the other, they encode information relevant to the property $p$.

\paragraph{Value-based Localization}
Given a target subnetwork $f_t(\cdot,  m_t \odot \theta_t)$ and a reference subnetwork $f_r(\cdot,  m_r \odot \theta_r)$, we select a subset of top-$k$ weights that are present in both subnetworks, but differ the most with regard to their values.
Let $I_{top}^k$ be the index set containing the indices of the $k$ largest absolute weight differences $|(m_r \odot \theta_r) -  (m_t \odot \theta_t)|$. Then we define the localization mask $b$ as: 
\begin{small}
\begin{equation}
     b_i = 
     \begin{cases}
        1,& \text{if } i \in I_{top}^k \\
        0,              & \text{otherwise}\,.
    \end{cases}
\end{equation}
\end{small}

\noindent We hypothesize that the weights with the largest absolute difference most strongly steer the networks towards opposing directions with respect to $p$.
\subsection{Contrastive Editing Strategies} \label{sec:edit_strategies}

After identifying subsets of weights potentially associated with the property of interest $p$, we use these weights to modify the target network. 
We explore different types of edits and evaluate their effectiveness.
In the following, we assume that we are given a target subnetwork $f_t(\cdot,  m_t \odot \theta_t)$, a reference subnetwork $f_r(\cdot,  m_r \odot \theta_r)$ and a localization mask $b$ that indicates which weights should be edited. The goal of each of the local editing strategies is to create a new target network $f_t'(\cdot,  \theta_t')$ that is modified with respect to the reference network.
 
\paragraph{Weight Interpolation (IP)}
Inspired by recent work on model merging~\citep{ilharco2022editing, wortsman2022model, yadav2024ties}, we propose linear weight interpolation that moves the localized weights of the target network closer to those of its reference or even adopts them completely~(${\alpha = 1}$):

\begin{small}
\begin{equation}
    \theta_t' = \theta_t + \alpha ((\theta_r - \theta_t) \odot b), \alpha \in [0, 1]\,.
\end{equation}
\end{small}

\noindent Note, that linear interpolation can also be used with mask-based localization by assuming that pruned weights have a value of 0.

\paragraph{Weight Extrapolation (EP)}
Similar to interpolation, we propose linear weight extrapolation to move the localized weights of the target either towards or away from those of the reference network:

\begin{small}
\begin{equation}
    \theta_t' = \theta_t + \alpha ((\theta_r - \theta_t) \odot b), \alpha \in \mathbb{R} \setminus [0, 1]\,.
\end{equation}
\end{small}

\noindent Allowing for weighting factors $\alpha \in \mathbb{R} \setminus [0, 1]$ enables flexible modifications, including e.g. the removal of a property from a network. 

\paragraph{Pruning (PR)}
Pruning is motivated by the assumption that the localized weights encode a property that can be eliminated by removing precisely those weights:

\begin{small}
\begin{equation}
    \theta_t' = \theta_t - (\theta_t \odot b)\,.
\end{equation}
\end{small}

\paragraph{Mask Switch (SW)}
Our final editing strategy is only applicable for mask-based localization and relies on the impact of weights being present (``turned on'') or pruned (``turned off'').
We apply the subnetwork mask of the reference model to the target model, resulting in pruning additional weights from the target model. Weights that were initially pruned in the target model during the localization step, but reinstated via the reference subnetwork mask, are restored to their values before pruning.

\begin{small}
\begin{equation}
    \theta_t' = \theta_t \odot m_r\,.
\end{equation}
\end{small}

\section{Experimental Setup}
We showcase the effectiveness of local contrastive editing in one of the, arguably, most established experimental environments for testing bias modification methods from the literature:  stereotypical binary gender bias encoded in the BERT \footnote{we use the Huggingface BERT-base-uncased distribution.} model~\citep{devlin-etal-2019-bert}. BERT is a widely used transformer model with 12 attention heads and 110 million parameters in total.

\subsection{Reference and Target Models} \label{sec_ref_target_models}

To localize and edit the encoding of stereotypical bias using contrastive strategies, we begin by establishing appropriate target and reference models. For obtaining a thorough understanding of the expected effects, we start from an ``extreme'' setup in which we intentially bias two types of models to be either \emph{stereotypical} or \emph{anti-stereotypical} concerning specific gender associations. This is accomplished by fine-tuning BERT on subsets of the English Wikipedia \footnote{\texttt{20220301.en} dump,~\citet{wikidump}} that we pre-processed to exhibit gendered associations using the well-established Counterfactual Data Augmentation~\citep{zhao-etal-2018-gender}. We describe the process in more detail in the following sections. \looseness=-1

\subsubsection{Bias Specification} \label{sec:bias_spec}
We investigate binary stereotypical gender bias in terms of stereotypical gender associations that manifest in written text.
We make use of an explicit bias specification $B = (T_1, T_2, A_1, A_2)$~\citep{doi:10.1126/science.aal4230, Lauscher_Glavaš_Ponzetto_Vulić_2020} that consists of two sets of target words $T_1, T_2$ that describe demographic groups between which we expect a bias w.r.t. two sets of attributes $A_1, A_2$. 
We choose terms in $T_1$ to represent the female gender (e.g.\ \emph{woman}) and terms in $T_2$ to describe the male gender (e.g. \emph{man}).  We then build pairs of corresponding terms $(t, t') \subset T_1 \times T_2$ (e.g.\ (\emph{aunt}, \emph{uncle})).
Further, we designate terms in $A_1$ to be stereotypically associated with $T_1$ (e.g. \emph{child-care}) and $A_2$ to contain attributes that are stereotypically associated with $T_2$ (e.g.\ \emph{programming}). The full specification can be found in appendix \ref{app:bias_spec} and was adopted from~\citet{barikeri-etal-2021-redditbias}. Note, that we do not claim our list of target and attribute words to be complete, we rather aim for a small and precise specification that demonstrates the feasibility of our approach. \looseness=-1

\subsubsection{Counterfactual Data Augmentation}

Starting from the bias specification in \ref{sec:bias_spec}, we create two datasets that we consider to be stereotypical and anti-stereotypical, respectively. Following the principle of Counterfactual Data Augmentation~\citep{zhao-etal-2018-gender}, we aim to artificially amplify or break associations between target words and their stereotypical attributes for obtaining our contrastive models. 
We use English Wikipedia as a base and filter the corpus for sentences $s_{(i,j)}$ containing exactly one target word $t \in T_i$ and one attribute word $a \in A_j$, where $i,j \in \{1, 2\}$. A sentence $s_{(i,j)}$ is categorized as stereotypical if $i = j$ and anti-stereotypical if $i \neq j$. For constructing a stereotypical dataset, we iterate through all sentences $s_{(i,j)}$ and retain those that are stereotypical. In cases where $s_{(i,j)}$ is anti-stereotypical, i.e. $i \neq j$, we replace the target term $t \in T_i$ with its corresponding paired term $t' \in T_j$. For creating an anti-stereotypical dataset, we keep all sentences $s_{(i,j)}$ with $i \neq j$ and substitute $t \in T_i$ with its paired target term $t' \in T_j$, if $i = j$. \looseness=-1
Note, that the resulting stereotypical and anti-stereotypical datasets are identical besides the swapped target terms. 

\subsubsection{Fine-tuning}
We fine-tune BERT on the biased datasets using a masked language modeling (MLM) objective, creating stereotypical and anti-stereotypical models. 
To achieve higher levels of bias, we adjust the masking function to mask the target and attribute terms from our bias specification  preferentially, i.e.\ with higher probability. We keep the average number of masked tokens constant by lowering the masking probability for all other tokens accordingly.
We tested preferential masking probabilities between $0.15$ and $0.9$ and found that a value of $0.3$ resulted in the best trade-off between perplexity and bias level.  
Additionally, we augmented the biased datasets with \emph{neutral} examples not containing terms from the bias specification as this positively impacted the stability of subnetworks. To ensure the robustness of our findings, we conduct our experiments using four different random seeds.
All training details can be found in appendix \ref{app:training_details}.

\subsection{Bias Evaluation} \label{sec:bias_eval}
We measure gender bias using three well-established bias benchmarks, namely WEAT, StereoSet and CrowS-Pairs, all measuring intrinsic bias.
The Word Embedding Association Test (WEAT)~\citep{doi:10.1126/science.aal4230} measures the differential association of two sets of target words w.r.t. two sets of attribute words based on embedding similarity. We utilize the WEAT 8 test that compares male and female target terms to attribute terms related to art and science, respectively (see appendix \ref{app_weat_8} for the full specification). As many of these terms are also included in our bias specification in \ref{sec:bias_spec}, WEAT 8 is a good instrument to precisely measure the gendered associations that we injected in our target and reference models and how these change when editing weights. We report the average effect size, where positive values indicate stereotypical bias and negative values indicate anti-stereotypical bias. We include a more detailed description of WEAT in appendix \ref{app:weat}.

We use the StereoSet intrasentence task~\citep{nadeem2020stereoset} as a second bias measure, evaluating stereotypical bias at the sentence level across four target domains. StereoSet contains sentences describing a target group with a blank (e.g. \emph{Girls tend to be more} \texttt{[BLANK]} \emph{than boys}), which can be filled with a stereotypical (e.g. \emph{soft}), an anti-stereotypical (e.g. \emph{determined}) or a meaningless (e.g. \emph{fish}) attribute. The \emph{StereoSet stereotype (SS) score}, ranging from 0 to 100, is computed based on how often a model fills the blank with a stereotypical attribute over an anti-stereotypical attribute. A SS score \textgreater 50 indicates stereotypical bias, while a score \textless 50 indicates anti-stereotypical bias. We report scores for the gender-related portion of the dataset. 

Our third measure, CrowS-Pairs~\citep{nangia2020crows}, comprises a crowd-sourced dataset of sentence pairs, with the first sentence being more stereotypical (e.g. \emph{Women are always too sensitive about things}) than the second (e.g. \emph{Men are always too sensitive about things}). The sentences differ only in the tokens related to the target group (e.g. \emph{women}/\emph{men}). 
Following~\citet{meade_2022_empirical} we calculate the \emph{CrowS-Pairs stereotype (CS) score} based on how often a model prefers the more stereotypical sentence. Scores range from 0 to 100, with CS scores \textgreater 50 indicating a preference for the more stereotypical sentence and a CS score of \textless 50 indicating the opposite. We report scores only for the gender-related portion of the dataset.

Both Stereoset and CrowS-Pairs have been criticized in the past for some of their properties~\citep{blodgett-etal-2021-stereotyping, pikuliak-etal-2023-depth, delobelle-etal-2022-measuring}. We use these measures only complementary to WEAT and will later show that we observe similar effects across all three measures.


\subsection{Subnetwork Identification} \label{sec:subnetworks}
We discover subnetworks through weight pruning, inspired by the lottery ticket hypothesis (LTH) \citep{frankle2018lottery}. The LTH posits that dense, randomly-initialized neural networks contain smaller subnetworks (\emph{winning tickets}) that can be trained in isolation to achieve the accuracy of the full model. We use  the approach of~\citet{chen2020lottery} and apply iterative magnitude pruning (IMP) to extract subnetworks from the stereotypical and anti-stereotypical models.
In IMP, we alternate between fine-tuning a model for $i$ steps and subsequently pruning $10\%$ of the weights with the lowest magnitude. After pruning, we reset the remaining weights to their initial values and repeat the steps until achieving a desired sparsity level.
We accept a subnetwork as winning ticket if its performance after $i$ fine-tuning steps is within $5\%$ of the performance of the full model, fine-tuned for the same number of steps. Additional details on our application of IMP can be found in appendix \ref{app:IMP}. Note that besides IMP, we also explored structured attention head pruning~\citep{prasanna-etal-2020-bert} but found no differences between stereotypical and anti-stereotypical subnetwork masks. We suspect that this occurred due to attention heads being too coarse-grained to capture the subtle differences in bias we injected.

\subsection{Uninformed Editing}
To explore the importance of the localization step, we also deploy our editing strategies to subsets of weights that are not informed by strategic localization. 
For \emph{uninformed intrapolation} and \emph{uninformed extrapolation}, we randomly sample from all weights, excluding those pruned in both the target and reference subnetworks, as intra- or extrapolation would not affect these weights. For \emph{uninformed pruning}, we randomly sample from the weights that are present in the target subnetwork, excluding those that already are pruned.
To ensure a fair comparison, we select the same number of weights as those identified by the localization strategies in section \ref{sec:localization}.

\section{Results} \label{sec:results}
We fine-tune BERT according to section \ref{sec_ref_target_models} and create models with stereotypical and anti-stereotypical biases. Table \ref{tab:bias} (``full'') shows that the stereotypical models exhibit higher levels of measurable stereotypical bias than the anti-stereotypical models. Thus, as intended, we have successfully steered the BERT model in two extreme directions, which will serve as a basis for our experiments on contrastive editing. We observe a trend for both types of models to shift towards the stereotypical regime, due to the fact that the bias benchmarks test for a broader range of associations than those we artificially controlled. This effect is less pronounced for WEAT, as WEAT specifically tests for many of our injected associations.

\begin{table}[!t]
  \small
  \centering
  \begin{subtable}[h]{\columnwidth}
  \centering
  \begin{tabular}{lrrrrr}
  \toprule
    & full & $10\%$ & $20\%$ & $30\%$ & $40\%$ \\
    \midrule
    WEAT        & 0.93 & 0.88 & 0.98 & 0.89 & 0.91 \\
    StereoSet   & 61.03 & 60.27 & 60.09 & 61.11 & 60.95 \\
    CrowS-Pairs & 57.92 & 59.45 & 59.83 & 57.25 & 58.88 \\
    \bottomrule
  \end{tabular}
  \caption{stereotypical subnetworks
  }
  \end{subtable}
    \newline
    \vspace{3mm}
    \newline
  \begin{subtable}[h]{\columnwidth}
  \centering
  \begin{tabular}{lrrrrr}
  \toprule
    & full & $10\%$ & $20\%$ & $30\%$ & $40\%$ \\
    \midrule
    WEAT        & -0.75 & -0.75 & -0.56 & -0.63 & -0.44 \\
    StereoSet   & 58.86 & 58.37 & 58.03 & 59.52 & 59.30 \\
    CrowS-Pairs & 52.77 & 53.15 & 52.86 & 52.01 & 52.39 \\
    
    \bottomrule
  \end{tabular}
  \caption{anti-stereotypical subnetworks
  }
    \end{subtable}
    \caption{\textbf{Bias of subnetworks at different sparsities}. We report the mean across all random seeds with higher scores indicating higher stereotypical bias.
  } \label{tab:bias}
\end{table}

\subsection{Subnetwork Analysis} \label{sec:subnet_analysis}
We discover subnetworks that are winning tickets (cf. section \ref{sec:subnetworks}) for both stereotypical and anti-stereotypical models at different sparsities up to $40\%$.
The discovered subnetworks are stable across runs with different random seeds, as indicated by a high Jaccard similaritiy of the pruning masks (\textgreater{} 0.98). 
This suggests that the findings are robust and not heavily influenced by factors such as specific data splits.
Table \ref{tab:bias} illustrates that the discovered subnetworks largely maintain the bias of their parent networks, highlighting their suitability for our contrastive approach. 

\begin{figure}[!t] 
\centering
\scalebox{0.8}{
  \includegraphics[width=\columnwidth]{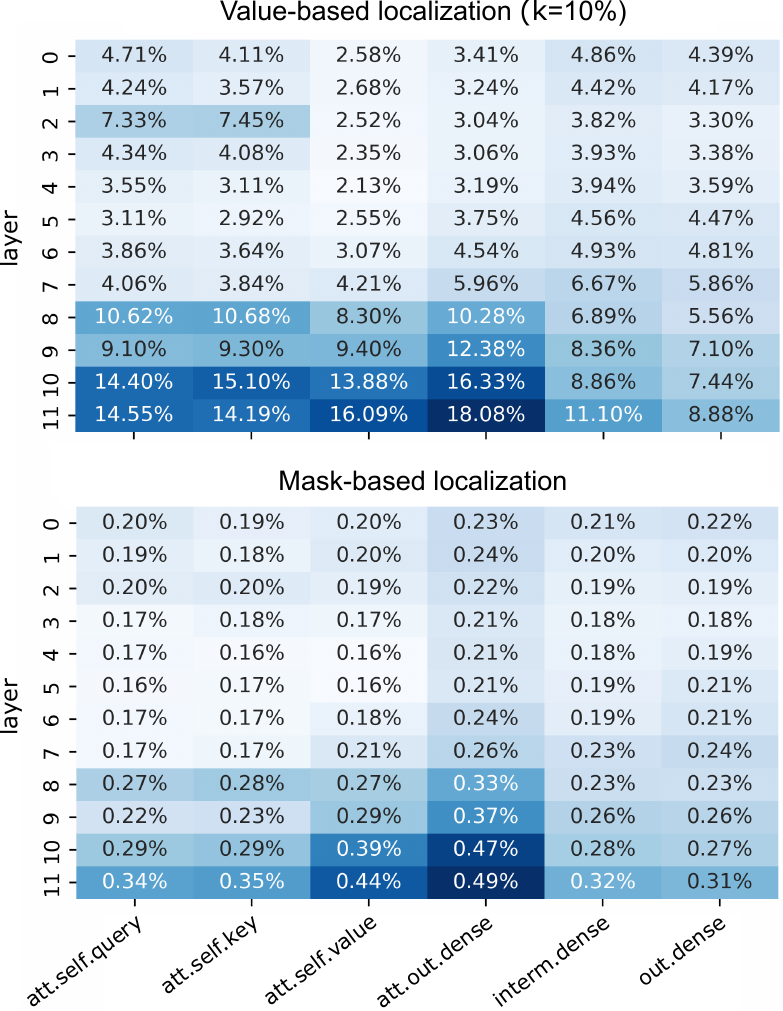}}
  \caption {\textbf{Bias localization.} We illustrate the percentage of weights per component that have been selected for editing. Notably, both localization strategies focus on the same layers and components. We show the results for subnetworks at sparsity $40\%$ and one random seed.} \label{fig:bias_localization}
\end{figure}

Next, we compare subnetworks with stereotypical bias to subnetworks with anti-stereotypical bias. At all sparsity levels, we find that the percentage of weights where the pruning masks differ remains below $0.5\%$, indicating a high degree of similarity. This is expected, as both types of subnetworks specialize on the same task and are fine-tuned on similar datasets, differing only in the injected stereotypical and anti-stereotypical associations, respectively.
To investigate where these associations manifest in the parameter space, we apply both mask-based and value-based localization. Although the localization strategies are designed not to select the same subsets of weights, they consistently target similar areas within the model, particularly focusing on the last layers and predominantly the attention output dense layer (see figure \ref{fig:bias_localization}). This observation aligns with previous work \citep{ma-etal-2023-deciphering, chintam-etal-2023-identifying}, which found that attention heads most influential for gender bias are located in higher layers, suggesting that bias is encoded in specific subcomponents of transformers.

\subsection{Effect on Gender Bias} \label{sec:eff_gender_bias}
\begin{figure*}[!t]
    \centering
    \includegraphics[width=0.9\linewidth]{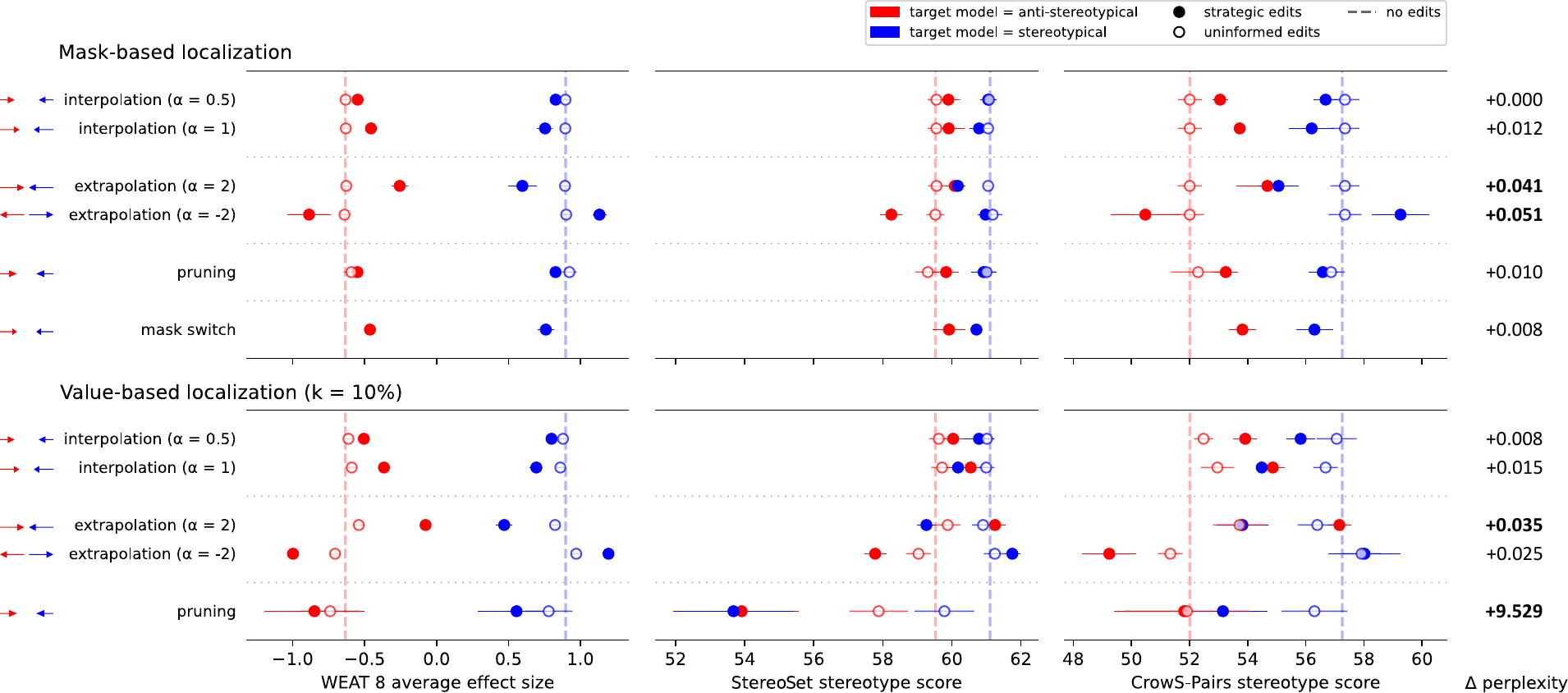} 
    \caption {\textbf{Local contrastive editing of gender bias.} We illustrate the effects of our local editing strategies on gender bias for settings in which (i) the target model is anti-stereotypical and the reference is stereotypical (\textcolor{red}{red}) and (ii) the target model is stereotypical and the reference is anti-stereotypical (\textcolor{blue}{blue}). The colored arrows on the left indicate the intuition of each strategy. We show the results for subnetworks at sparsity 30\% and report the mean bias over four random seeds, with error bars indicating one standard deviation in each direction. On the right, we display the mean perplexity change across both settings and all random seeds where bold indicates a significant increase. Our local editing strategies can successfully steer stereotypical bias with both localization methods, while uninformed edits have much lower or no effect at all. Results for other sparsities can be found in appendix \ref{app:add_main}.} \label{fig:main_result}
    \end{figure*}

We investigate the effectiveness of the local contrastive editing strategies by considering two settings: using the stereotypical subnetworks as the target with the anti-stereotypical subnetworks as the reference, and vice versa. Figure \ref{fig:main_result} illustrates results for selected parameter settings, demonstrating the flexibility of our strategies.
    
We find that nearly all editing strategies, when combined with either mask- or value-based localization, effectively modify gender bias as intended. Mask-based editing achieves this efficiently with a small subset size of less than $0.5\%$. By varying the weighting factor $\alpha$, we can flexibly control the bias of the target model and, according to WEAT, even completely remove bias at $\alpha=2$. In contrast, uninformed edits result in minimal or no changes to the bias scores, highlighting the critical role of the localization step in local contrastive editing.
We summarize that (i) both mask-based and value-based localization can identify subsets of weights driving stereotypical gender bias, and that (ii) gender bias can be controlled through contrastive editing strategies on these subsets.

\subsection{Effect on Performance}
We further examine how local contrastive editing affects a model's ability to model language. To assess this, we use perplexity as a primary measure and additionally compute the \emph{language modeling (LM) score} as introduced by~\citet{nadeem2020stereoset}. Similar to the SS score (cf. section \ref{sec:bias_eval}), the LM score is computed on the StereoSet dataset and measures how frequently a model prefers a sentence with a meaningful association (e.g. \emph{Girls tend to be more \textbf{determined} than boys}) over a nonsensical one (e.g. \emph{Girls tend to be more \textbf{fish} than boys}). The LM score ranges from 0 to 100, where 100 represents an ideal model that always favours the semantically meaningful association.

As shown in figure \ref{fig:main_result}, nearly all editing strategies lead to only minor increases in perplexity. The notable exception is value-based pruning, which causes a significant increase of 9.529 points in perplexity, while extrapolation also leads to a significant but much smaller rise. This substantial degradation in model performance may explain the counterintuitive effect of value-based pruning on gender bias in figure \ref{fig:main_result}, as the models' overall functionality is severely impaired. 
Consistent with these findings, the LM score shows no significant decline, except for value-based pruning, which leads to a significant drop of 2.723 points at a sparsity level of 30\%.
Complete LM score results and results for other sparsity levels can be found in appendix \ref{app:add_LM}.

\begin{table}[!t]
  \small
  \centering
  \begin{minipage}[t]{0.48\columnwidth}
    \centering
    \resizebox{\columnwidth}{!}{
      \begin{tabular}{l@{\hspace{3pt}}rr}
        \toprule
        \textbf{Model} & MNLI-m/mm & STS-B \\
        \midrule
        \textbf{Base} & \textbf{84.4/84.8} & \textbf{89.0/88.9} \\
        \midrule
        IP ($\alpha$=0.5) & 83.9/84.2 & 88.6/88.0\\
        IP ($\alpha$=1) & 83.9/84.4 & 88.5/88.0\\
        EP ($\alpha$=2) & 83.9/84.3 & 88.5/88.0\\
        EP ($\alpha$=-2) & 83.8/84.4 & 88.5/88.1\\
        PR & 83.9/84.3 &  88.4/87.9\\
        SW & 83.8/84.4 & 88.5/88.0  \\
        \bottomrule
      \end{tabular}
    }
    \subcaption{mask-based loc.}
  \end{minipage}
  \hfill
  \begin{minipage}[t]{0.48\columnwidth}
    \centering
    \resizebox{\columnwidth}{!}{
      \begin{tabular}{l@{\hspace{3pt}}rr}
      \toprule
        \textbf{Model} & MNLI-m/mm & STS-B \\
        \midrule
        \textbf{Base} & \textbf{84.4/84.8} & \textbf{89.0/88.9} \\
        \midrule
        IP ($\alpha$=0.5) & 83.9/84.3 & 88.6/88.0\\
        IP ($\alpha$=1) & 83.8/84.3 & 88.5/88.0\\
        EP ($\alpha$=2) & 84.0/84.4 & 88.5/87.9\\
        EP ($\alpha$=-2) & 83.8/84.5 & 88.6/88.1\\
        PR & 83.5/83.8 &  87.0/86.6 \\
        &&  \\
        \bottomrule
      \end{tabular}
    }
    \subcaption{value-based loc. ($k$=$10\%$)}
  \end{minipage}
  \caption{\textbf{Performance on downstream tasks}. We fine-tune the edited models on the MNLI and STS-B tasks from the GLUE benchmark and show the results for the stereotypical target model at 30\% sparsity using a single random seed. For MNLI, we report both matched and mismatched accuracy while for STS-B, we present Pearson and Spearman correlation. Results for other sparsities and target models can be found in the appendix \ref{app:add_downstream}.}
  \label{tab:glue}
\end{table}

Next, we fine-tune the edited models on the MNLI and STS-B tasks from the GLUE benchmark~\citep{wang-etal-2018-glue}, to evaluate their performance on downstream tasks. As shown in table \ref{tab:glue}, the model subjected to value-based pruning again exhibits the most significant performance drop compared to the base model~\footnote{BERT-base-uncased}. Models edited using other strategies experience a maximum performance loss of only $0.71\%/0.59\%$ for MNLI and  $1.07\%/1.06\%$ for STS-B.

In summary, we demonstrate that, with exception of value-based pruning, local contrastive editing largely preserves a model’s language modeling ability. Furthermore, the edited models can still be effectively used in downstream tasks with only minor performance loss. 

\subsection{Ablation Study}
We explore the impact of the number of selected weights $k$, the weighting factor $\alpha$, and the different localization strategies on inter- and extrapolation.

\paragraph{Number of Selected Weights}

\begin{figure*}[!t]
\centering
  \includegraphics[width=0.75\linewidth]{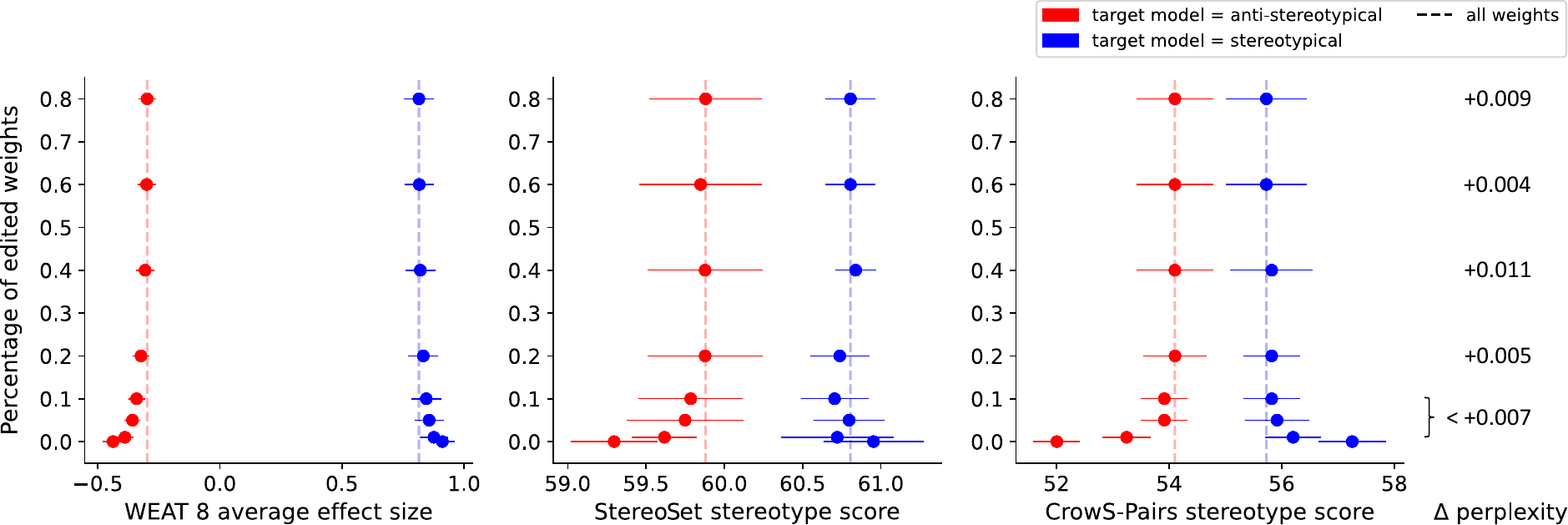}
  \caption {\textbf{Sensitivity to the number of weights edited.} 
  We explore the influence of the number of top-$k$ weights that are used for value-based interpolation with ${\alpha =0.5}$. We show the results for subnetworks at sparsity $30\%$ and report the mean bias and standard deviation across four random seeds. We report the average change in perplexity across both target models and all random seeds, observing no significant increase for any choice of $k$. Results for other sparsities can be found in appendix \ref{app:add_topk}.} \label{fig:sensitivity_k}
\end{figure*}

Figure \ref{fig:sensitivity_k} shows the sensitivity of value-based interpolation ($\alpha =0.5$) to the number of selected weights $k$. Increasing $k$ initially leads to stronger effects on gender bias up to a threshold of 20\textendash40\%. Beyond this range, editing further weights seems to have no effect, indicating that there is a critical subset primarily encoding the bias. \looseness=-1

\paragraph{Weighting Factor}

\begin{figure*}[!t]
    \centering
    \begin{subfigure}[b]{0.7\linewidth}
        \centering
        \includegraphics[width=\linewidth]{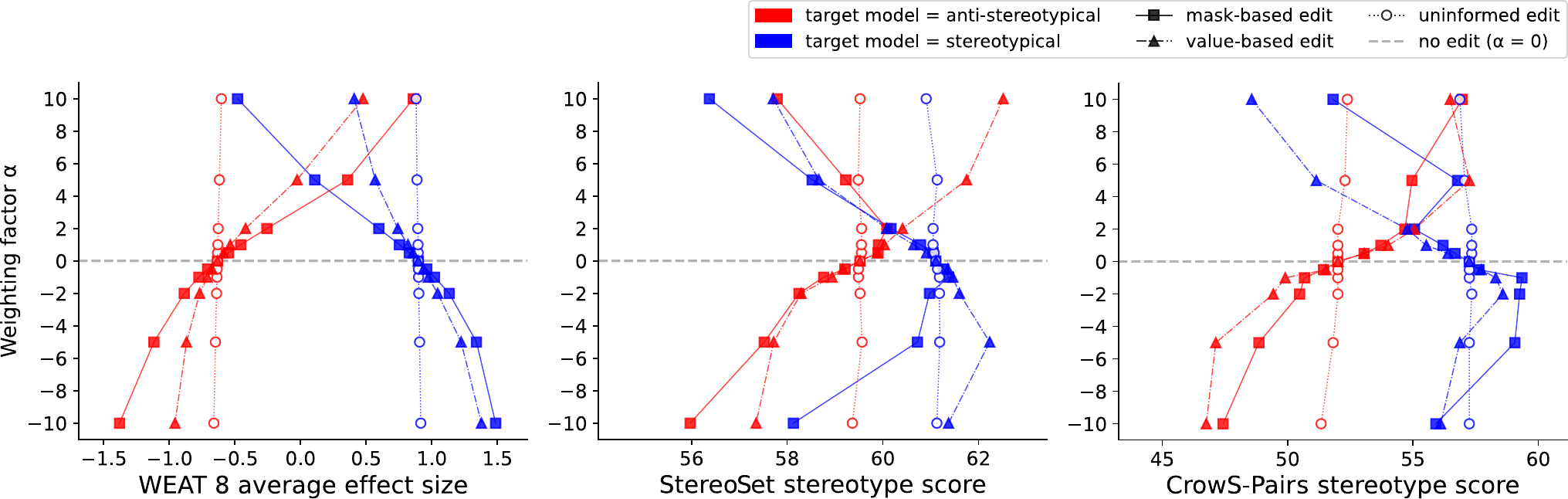}
        \caption {Gender bias}
        \label{fig:alpha_bias}
    \end{subfigure}
    \hfill
    \begin{subfigure}[b]{0.24\linewidth}
        \centering
        \includegraphics[width=\linewidth]{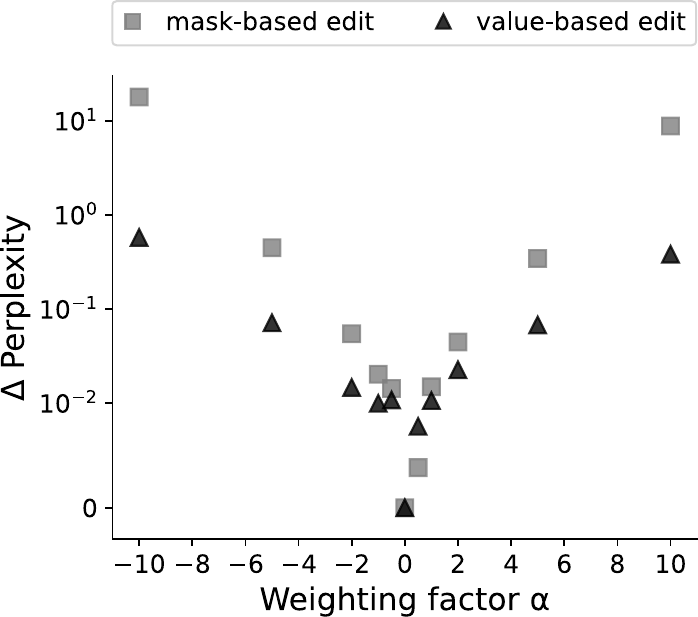}
        \caption {Perplexity}
        \label{fig:alpha_perplexity}
    \end{subfigure}
    \caption {\textbf{Sensitivity to the weighting factor.} We investigate the effect of different weighting factors $\alpha$ on gender bias (a) and perplexity (b). For value-based and uninformed edits we choose the same number of weights that were selected by mask-based localization. We find that weighting factors with higher magnitudes lead to greater effects on bias, correlating with an increase in perplexity. We display the results for a sparsity level of 30\% and report the mean across four random seeds.  For perplexity, we average the results for both target models. Results for other sparsities can be found in appendix \ref{app:add_alpha}.} 
    \label{fig:alpha_sensitivity}
\end{figure*}

Figure \ref{fig:alpha_sensitivity} shows the behavior of linear weight inter- and extrapolation for all localization strategies. For WEAT, we can achieve a smooth, monotonous change in gender bias by gradually increasing $\alpha$. 
StereoSet and CrowS-Pairs measure a similar trend, but the effects become inconsistent for higher absolute weighting factors ($|\alpha| \geq 5$), likely due to a decline in language modeling performance, as evidenced by increasing perplexity (see figures \ref{fig:alpha_perplexity}, \ref{fig:alpha_perplexity_all_sp}) and decreasing LM scores (see figure \ref{fig:alpha_LM_all_sp}). 
Overall, we find that varying $\alpha$ allows flexible control and calibration of bias levels within a target model, that can be tailored to the characteristics of the reference model (e.g. reducing bias when the reference model itself exhibits bias), albeit within certain limits.

\paragraph{Localization Strategies}
We choose the same number of weights to be edited for all localization strategies, allowing their direct comparison in figure \ref{fig:alpha_sensitivity}.
We observe that the localization strategy that leads to stronger steering effects as measured by WEAT (specifically mask-based localization at 30\% sparsity) also leads to a stronger decline in language modeling ability when $|\alpha|$ increases.  
In line with the result in figure \ref{fig:main_result}, we further find that uninformed editing does not change the bias level significantly, not even for high magnitudes of $\alpha$.
This suggests that certain subsets of weights encode gender bias more prominently, and that their localization can be crucial for bias modification.

\subsection{Wider Applicability}
So far, our experiments have been conducted in a controlled environment where the target and reference models were fine-tuned on parallel datasets. To test the wider applicability of our approach, we fine-tune a \emph{neutral} model on a subset of Wikipedia that is independent of the biased datasets described in section \ref{sec_ref_target_models}.
The training details can be found in appendix \ref{app:training_details}. In line with the other experiments, we extract neutral subnetworks at different sparsities with four random seeds. We use those as target networks and edit them w.r.t both stereotypical and anti-stereotypical reference models. 
\begin{figure}[!t]
  \includegraphics[width=\linewidth]{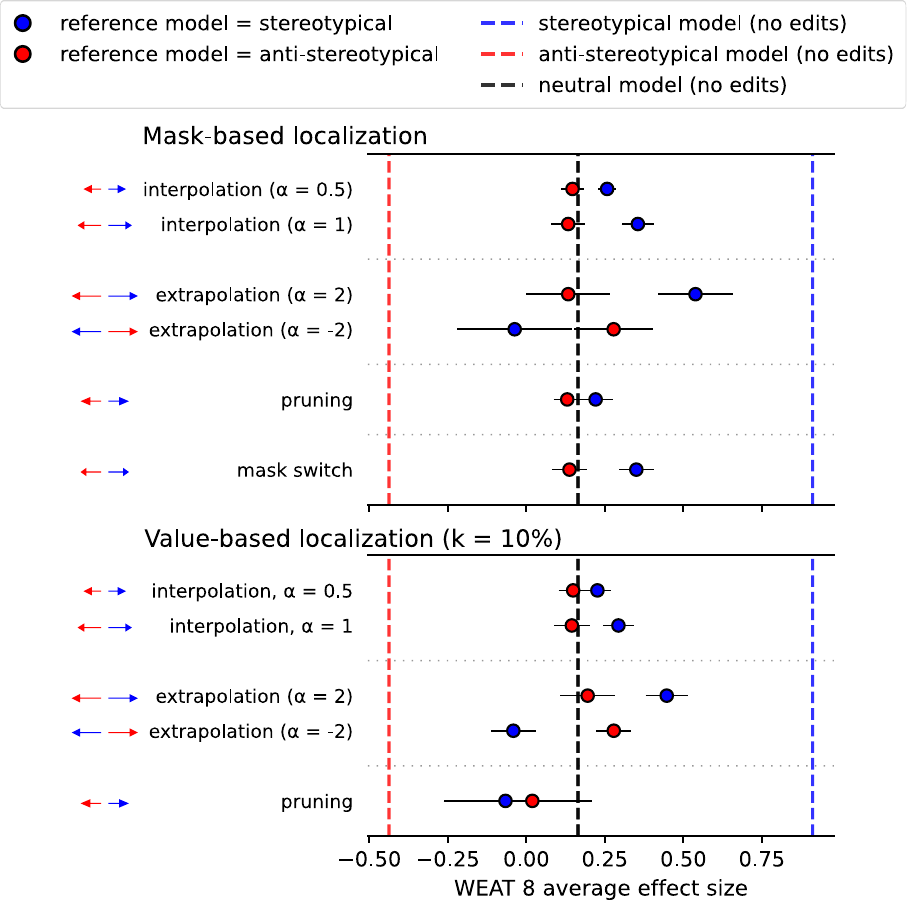} 
  \caption {\textbf{Application to a neutral model.} We apply local contrastive editing to a \emph{neutral} target model. Using a stereotypical reference model (\textcolor{blue}{blue}) can effectively steer the neutral model's bias. Using an anti-stereotypical reference model (\textcolor{red}{red}) produces inconclusive results. We show the results for a sparsity of $40\%$ and report the mean and standard deviation across four random seeds. Results for other sparsities can be found in appendix \ref{app:add_neutral}} \label{fig:neutral_both}
\end{figure}
Figure \ref{fig:neutral_both} illustrates the results at sparsity $40\%$. By using the stereotypical reference model we can successfully modify the bias of the neutral model in line with our intuition. 
For instance, extrapolation with ${\alpha = -2}$, successfully removes stereotypical bias, as measured by WEAT. 
This is not trivial, as here the reference and target models are fine-tuned on datasets that do not overlap, which implies that the weights selected by the localization strategies may encode differences in the datasets beyond just stereotypical bias.
Using an anti-stereotypical reference model produces mixed results, aligning with expectations in most but not all scenarios, requiring further investigation.

\section{Conclusion}

Our research shows that stereotypical gender bias is primarily encoded in specific subsets of weights within LMs. We propose various local contrastive editing strategies and demonstrate that they can effectively identify and modify these subsets to flexibly control and mitigate gender bias.
This work enhances our understanding of where stereotypical biases manifest in the parameter space of LMs and opens up new avenues for developing parameter-efficient strategies for model editing in a contrastive manner. 
Local contrastive editing is not limited to gender bias, and future research could explore its application to other tasks and domains.


\section{Limitations}
Naturally, our work comes with limitations. We conduct experiments using a single model architecture and a single bias type only. We restrict our study to this model architecture because of computational constraints and environmental considerations, particularly because iterative magnitude pruning requires substantial computational resources. However, we anticipate that our findings generalize broadly, as related work on weight averaging has been shown to generalize to other model architectures as well~\citep{wortsman2022model, yadav2024ties, ilharco2022patching}. Moreover, findings from other studies on bias mitigation suggest generalizability to other types of bias that share similar specifications~\citep{hauzenberger-etal-2023-modular, guo-etal-2022-auto}.


\section{Ethical Considerations}
While our work ultimately targets the development of  strategies for reducing bias in language models, it is important to acknowledge the potential dual use of these techniques.
The same strategies designed to reduce bias, can also be used to perpetuate and amplify biases in language models. 
Moreover, our intentional design of language models that initially exhibit high levels of bias raises concerns about their potential deployment in various applications. Despite our intention to mitigate bias, the existence of such models risks normalizing and perpetuating stereotypical gender biases in society. 

Additionally, our study focuses on bias specifically between male and female groups, thus operating on a binary specification of gender bias, only. We recognize that gender is a spectrum, encompassing identities beyond strictly male and female categories. 
Our rationale for this approach is to evaluate the effectiveness of our strategies using an existing and well-established specification that facilitates measurement of their effects.

\section*{Acknowledgments}
We thank Ekaterina Shutova for her valuable input and feedback for this project.
The authors acknowledge support by the state of Baden-Württemberg through bwHPC and the German Research Foundation (DFG) through grant INST 35/1597-1 FUGG. 
The work of Anne Lauscher is funded under the Excellence Strategy of the German Federal Government and the Federal States. The work of Rochelle Choenni is supported by a Google PhD Fellowship.

\bibliography{anthology,custom}

\appendix

\section{Gender Bias Specification}
\label{app:bias_spec}
\paragraph{Target Word Pairs} \mbox{} \\
$T_1 \times T_2$: \emph{(woman, man), (women, men), (girl, boy), (mother, father), (daughter, son), (wife, husband), (niece, nephew), (mom, dad), (bride, groom), (lady, gentleman), (madam, sir), (hostess, host), (female, male), (wife, husband), (aunt, uncle), (sister, brother), (she, he)}

\paragraph{Attribute Words} \mbox{} \\
$A_1$: \emph{nurse, secretary, housekeeper, cook, sewing, cleaning, waitress, maid, child-care, receptionist, clerk, home-care, poetry, art, dance, literature, novel, symphony, drama, sculpture, shakespeare} \\
$A_2$: \emph{surgeon, executive, manager, officer, engineering, programming, lawyer, engineer, finance, administrator, physician, science, math, geometry, technology, equation, computation, physics, chemistry, einstein}

\section{Training Details} \label{app:training_details}

For fine-tuning and pruning we used 4 NVIDIA A100-80GB GPUs. One pruning and fine-tuning iteration took 32 GPU hours, amounting to 160 GPU hours to extract subnetworks up to sparsity $40\%$ from a single model. As we repeated this for three types of models and four random seeds, we amount in a total of 1920 GPU hours.

\begin{table}[h]
    \centering
    \begin{tabular}{lll}
    \toprule
               & \textbf{Biased} & \textbf{Neutral}\\
    \midrule
    \# Biased examples      &  164,524 & 0           \\
    \# Neutral examples     & 164,524 & 329,048        \\
    Task                    & MLM & MLM \\
    Masking prob. &    0.3 / 0.148 & 0.15                    \\
    \# Epochs               & 3 & 3        \\
    \# Iterations/epoch     & 4627 & 4627   \\
    Batch size              &  64 & 64        \\
    Learning rate           &  $1 \times 10^{-5}$   & $1 \times 10^{-5}$      \\
    Eval size & 0.1 & 0.1 \\
    Eval Measure & Perplexity & Perplexity \\
    \# Random seeds & 4 & 4 \\
    \bottomrule
    \end{tabular}
    \caption{Details of fine-tuning biased (stereotypical or anti-stereotypical) and neutral models. Optimization is performed with AdamW with $\epsilon = 1 \times 10^{-8}$ and the learning rate decays linearly to zero. We use standard implementations and hyperparameter settings~\citep{wolf-etal-2020-transformers}.   
    }
\end{table}

\section{Word Embedding Association Test (WEAT)} \label{app:weat}
\citet{doi:10.1126/science.aal4230} introduce WEAT by extending the Implicit Association Test~\citep{greenwald1998measuring}, a test used to measure human biases, to word embeddings. The test measures the differential association of two sets of target words $X, Y$ (e.g. \emph{female} and \emph{male} terms) w.r.t. two sets of attribute words $A, B$ (e.g. \emph{art} and \emph{science} terms) based on their cosine similarity in the embedding space:

\begin{small}
\begin{equation*}
    s(X,Y,A,B) = \sum_{x\in X} s(x, A, B) - \sum_{y\in Y} s(y, A, B)\,,
\end{equation*}
\end{small}

\noindent where

\begin{small}
\begin{equation*}
    s(w, A, B) = \frac{1}{|A|} \sum_{a \in A} cos(w, a) - \frac{1}{|B|} \sum_{b \in B} cos(w, b)\,.
\end{equation*}
\end{small}

\noindent The significance of the test is computed with a permutation test, where $\{(X_i, Y_i)\}_i$ are all equally sized partitions of $X \cup Y$ into two sets:
\begin{equation*}
    Pr_i((s(X_i, Y_i, A, B) > s(X,Y,A,B))
\end{equation*}
Here, we report the effect size as a measure of separation between the association distributions:
\begin{equation*}
    \frac{\text{mean}_{x \in X} s(x, A, B) - \text{mean}_{y \in Y} s(y, A, B)}{\text{std-dev}_{w \in X \cup Y} s(w,A,B)}
\end{equation*}
 Following~\citet{vulic2020multi}, we extract embeddings for all target and attribute words by feeding them through BERT, prepended with the start of sequence token and appended with the separator token (e.g. \texttt{[CLS] woman [SEP]}). We then extract embeddings from each hidden layer and compute WEAT separately for each layer. Finally, we report the average effect size of the test across all layers. 

\section{WEAT 8 Specification} \label{app_weat_8}

\paragraph{Target Words} \mbox{} \\
$X$: \emph{science, technology, physics, chemistry, Einstein, NASA, experiment, astronomy} \\
$Y$: \emph{poetry, art, Shakespeare, dance, literature, novel, symphony, drama}

\paragraph{Attribute Words} \mbox{} \\
$A$: \emph{brother, father, uncle, grandfather, son, he, his, him} \\
$B$: \emph{sister, mother, aunt, grandmother, daughter, she, hers, her}

\section{Iterative Magnitude Pruning} \label{app:IMP}
We apply iterative magnitude pruning according to the following procedure:
\begin{enumerate}
    \item Fine-tune a pre-trained network $f(\cdot, \theta_0)$ for $i$ steps
    \item Globally prune $p\%$ of the weights with the lowest magnitude, resulting in a subnetwork $f(\cdot, m \odot \theta_i)$ with pruning mask $m$
    \item Reset the remaining weights to their initial values $\theta_0$
    \item Repeat the previous steps on $f(\cdot, m \odot \theta_0)$ until the desired sparsity level is reached
\end{enumerate}
We set the pruning rate per iteration to ${p\% = 10 \%}$.
Consistent with~\citet{chen2020lottery}, we only prune weights (and e.g. not biases) and exclude the embedding layer and the task-specific layer from the pruning process.
In each fine-tuning iteration, we use the number of steps and parameter settings detailed in appendix \ref{app:training_details}.

\section{Additional Results}

\subsection{Effect of Local Contrastive Editing on Gender Bias} \label{app:add_main}

\begin{figure*}[h]
\centering
  \includegraphics[width=0.8\textwidth]{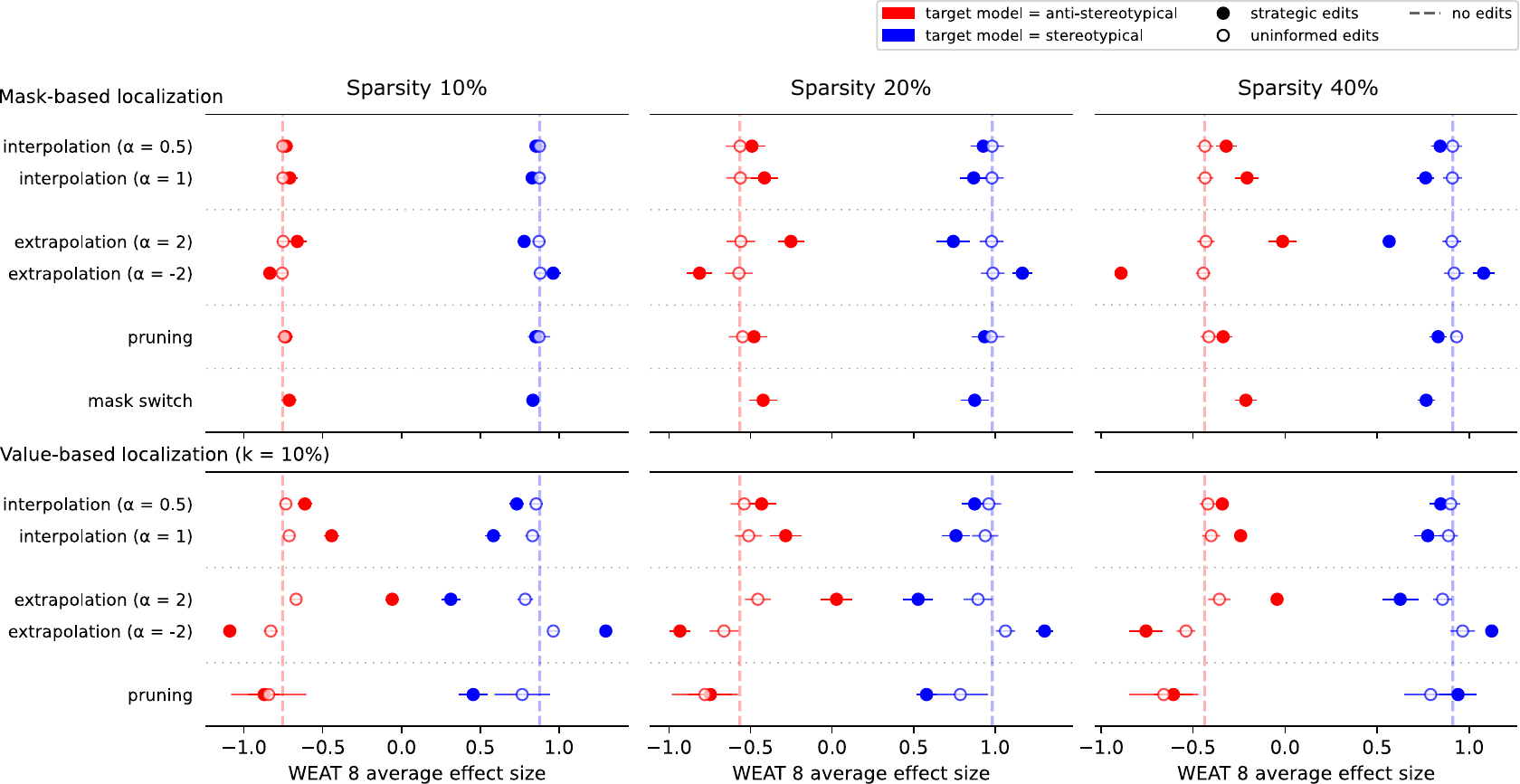} 
  \caption {\textbf{WEAT average effect size after local contrastive editing.} We report the mean bias at different sparsity levels across four random seeds with error bars indicating one standard deviation in each direction.} \label{fig:main_weat_all_sp}
\end{figure*}

\begin{figure*}[h]
\centering
  \includegraphics[width=0.85\textwidth]{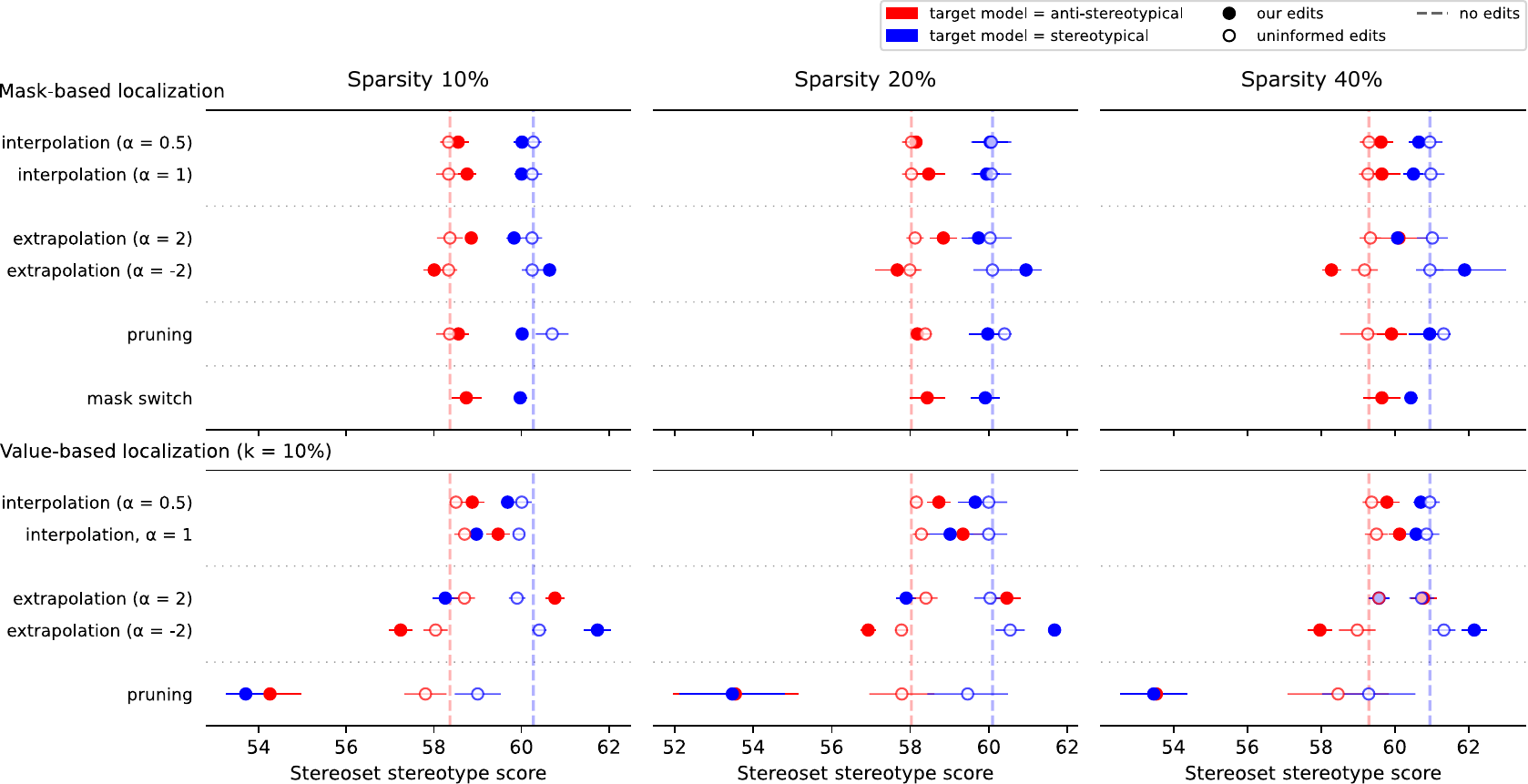} 
  \caption {\textbf{StereoSet stereotype scores after local contrastive editing.} We illustrate the mean bias at different sparsity levels across four random seeds with error bars indicating one standard deviation in each direction.} \label{fig:main_stereoset_all_sp}
\end{figure*}

\begin{figure*}[h]
\centering
  \includegraphics[width=0.85\textwidth]{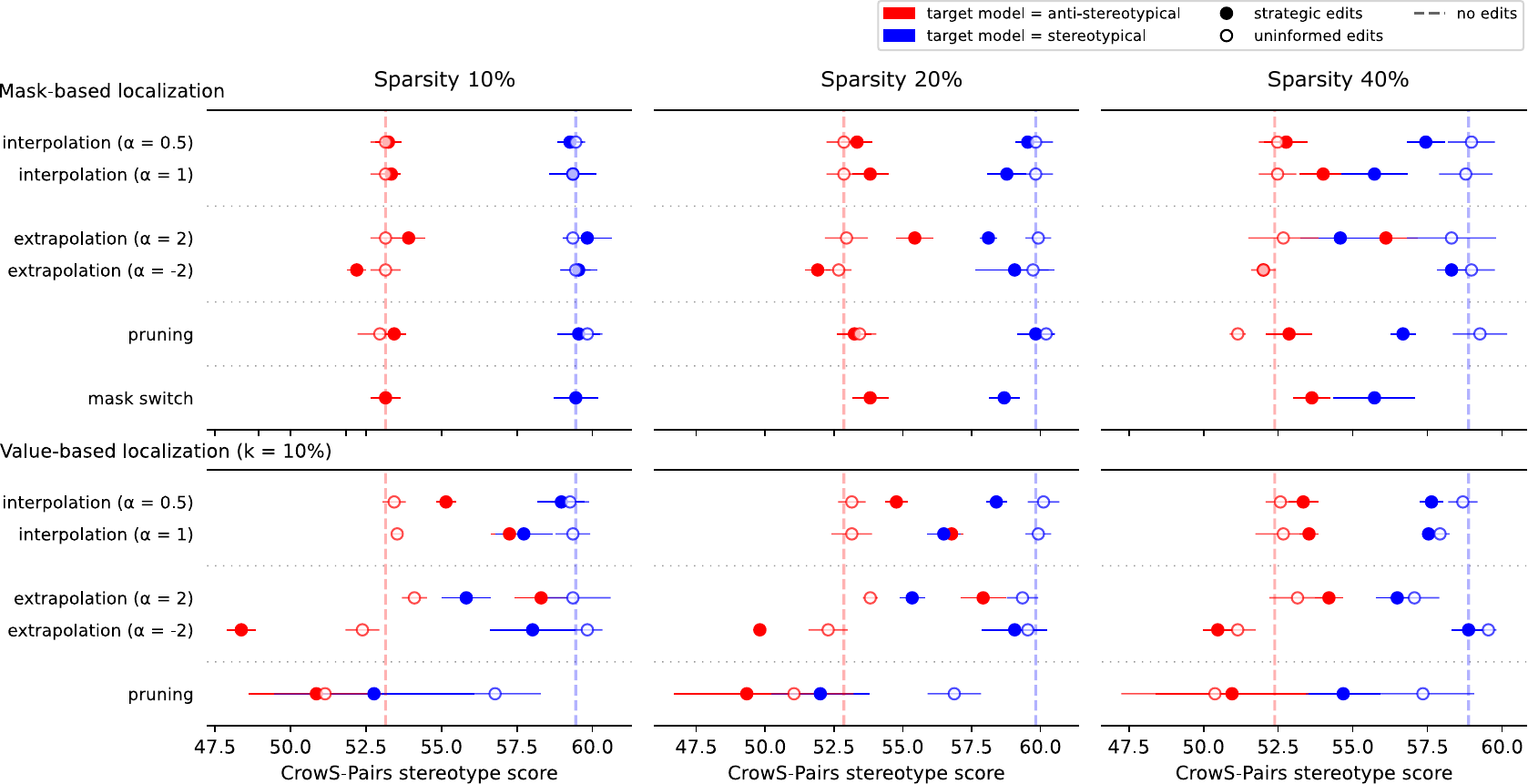} 
  \caption {\textbf{CrowS-Pairs stereotype scores after local contrastive editing.} We show the mean bias at different sparsity levels across four random seeds with error bars indicating one standard deviation in each direction.} \label{fig:main_crows_all_sp}
\end{figure*}

We show the effects of local contrastive editing on gender bias for additional sparsity levels in figures \ref{fig:main_weat_all_sp}, \ref{fig:main_stereoset_all_sp} and \ref{fig:main_crows_all_sp}. We observe that as the sparsity level increases, the impact of mask-based editing becomes more pronounced, whereas the influence of value-based editing on bias declines.

\subsection{Language Modeling Ability} \label{app:add_LM}
We present the language modeling ability (as measured by perplexity and LM score) of the discovered subnetworks at varying sparsity levels in table \ref{tab:app_subn_LM}. Both, perplexity and LM score are comparable between stereotypical and anti-stereotypical subnetworks and remain similar across different sparsity levels.

Table \ref{tab:app_perf} presents the change in language modeling ability after local contrastive editing. We present the average changes in perplexity and LM score across both target models and four random seeds at different sparsity levels. We apply a one-sided Wilcoxon signed-rank test with Bonferroni correction to assess whether the observed changes (an increase in perplexity or a decrease in LM scores) are significant. Notably, across all sparsity levels, perplexity remains consistently low, with significant increases occurring occasionally after extrapolation with large absolute weighting factors, and consistently across all sparsities for value-based pruning. A similar trend is observed for LM score, with significant drops occurring only in the case of value-based pruning.

\begin{table}[h]
  \small
  \centering
  \begin{subtable}[h]{\columnwidth}
  \centering
  \begin{tabular}{lrrrrr}
  \toprule
    & full & $10\%$ & $20\%$ & $30\%$ & $40\%$ \\
    \midrule
    Perplexity & 5.57 & 5.56 & 5.59 & 5.69 & 5.88 \\
    LM score   & 85.58 & 85.51 & 85.80 & 85.81 & 85.53 \\
    \bottomrule
  \end{tabular}
  \caption{stereotypical subnetworks
  }
  \end{subtable}
    \newline
    \vspace{3mm}
    \newline
  \begin{subtable}[h]{\columnwidth}
  \centering
  \begin{tabular}{lrrrrr}
  \toprule
    & full & $10\%$ & $20\%$ & $30\%$ & $40\%$ \\
    \midrule
    Perplexity  & 5.57 & 5.58 & 5.60 & 5.68 & 5.90 \\
    LM score & 85.88 & 85.63 & 85.89 & 85.78 & 85.51 \\
    
    \bottomrule
  \end{tabular}
  \caption{anti-stereotypical subnetworks
  }
    \end{subtable}
    \caption{\textbf{Language modeling ability of subnetworks at different sparsities}. We report the mean across all random seeds, where \emph{lower} perplexity and \emph{higher} LM scores indicate better performance in terms of language modeling.
  } \label{tab:app_subn_LM}
\end{table}

\begin{table*}[t]
  \small
  \centering
    \begin{tabular}{lrrrrrrrr}
    \toprule
           & \multicolumn{4}{c}{$\Delta$ perplexity $\downarrow$} & \multicolumn{4}{c}{$\Delta$ LM score $\uparrow$} \\
           \cmidrule(lr){2-5} \cmidrule(lr){6-9}
           &     10\% &   20\%  &   30\% &    40\%  &  10\% &      20\% &   30\%  &  40\%\\
        \midrule
        Mask-based localization &&&& \\
        IP ($\alpha=0.5$) &     -0.002 &   +0.001 &   +0.000            &   +0.007          & +0.012 & -0.010 & +0.012 & -0.009 \\
        IP ($\alpha=1$)   &     -0.003 &   +0.001 &   +0.012            &   \textbf{+0.017} & -0.080 & -0.045 & -0.080 & -0.038 \\
        EP ($\alpha=2$)   &     +0.006 &   +0.014 &   \textbf{+0.041}   &   \textbf{+0.072} & +0.005 & -0.016 & +0.005 & +0.015 \\
        EP ($\alpha=-2$)  &     +0.002 &   +0.016 &   \textbf{+0.051}   &   \textbf{+0.137} & -0.023 & -0.050 & -0.023 & -0.099 \\
        PR                &     -0.003 &   -0.001 &   +0.010            &   +0.014          & -0.005 & -0.023 & -0.005 & -0.017 \\
        SW                &     +0.000 &   +0.010 &   +0.008            &   +0.019          & -0.055 & -0.061 & -0.055 & -0.023 \\
        \midrule
        Value-based localization ($k=10\%$) &&&& \\
        IP ($\alpha=0.5$)      &   -0.003           &   +0.006          &   +0.008          &   +0.002          & -0.004          & +0.013          & -0.004          & -0.010 \\
        IP ($\alpha=1$)        &   +0.010           &   +0.013          &   +0.015          &   \textbf{+0.016} & -0.076          & -0.017          & -0.076          & +0.005 \\
        EP ($\alpha=2$)        &   \textbf{+0.051}  &   \textbf{+0.032} &   \textbf{+0.035} &   \textbf{+0.032} & -0.147          & -0.048          & -0.147          & +0.020 \\
        EP ($\alpha=-2$)       &   +0.031           &   +0.020          &   +0.025          &   +0.017          & +0.005          & -0.233          & +0.005          & +0.059 \\
        PR                     &  \textbf{+9.225}   &  \textbf{+10.18}  &  \textbf{+9.529}  &  \textbf{+11.722} & \textbf{-2.724} & \textbf{-3.500} & \textbf{-2.723} & \textbf{-0.965} \\
        \bottomrule
        \end{tabular}
  \caption{\textbf{Change in language modeling ability.} We show the mean change in perplexity and LM score after local contrastive editing across both target models (stereotypical and anti-stereotypical) and four random seeds at different sparsity levels. We print significant differences bold.}
  \label{tab:app_perf}
\end{table*}

\subsection{Downstream Tasks} \label{app:add_downstream}
Tables \ref{tab:app_glue_stereo} and \ref{tab:app_glue_anti} present the results of fine-tuning the edited models on the MNLI and STS-B tasks from the GLUE benchmark for different sparsity levels. 
The edited models exhibit only a slight decrease in performance compared to the base model, with the lowest performing models achieving $82.9/83.4$ ($- 1.8\%/-1.2\%$) on MNLI and $86.6/86.24$ ($-2.7\%/-3.0\%$) on STS-B. 
We note that at each sparsity level, the models with the greatest performance drop were those subjected to value-based pruning prior to fine-tuning on the downstream task.

\begin{table*}[t]
  \small
  \centering
  \resizebox{0.8\textwidth}{!}{
    \begin{tabular}{lrrrrrr}
    \toprule
           & \multicolumn{3}{c}{MNLI-m/mm $\uparrow$} & \multicolumn{3}{c}{STS-B $\uparrow$}\\
           \cmidrule(lr){2-4} \cmidrule(lr){5-7}
           &     10\% &   20\%  &   40\%  &  10\% &   20\%  &    40\% \\
        \midrule
        \textbf{Base} & \multicolumn{3}{c}{\textbf{84.4/84.8}} & \multicolumn{3}{c}{\textbf{89.0/88.9}} \\
        \midrule
        Mask-based loc. &&&&&& \\
        IP ($\alpha=0.5$) &  84.6/84.5 &  84.4/84.6 &  84.1/84.0 &  88.7/88.3 &  88.7/88.4 &  88.2/87.6\\
        IP ($\alpha=1$)   &  84.5/84.5 &  84.3/84.5 &  84.0/84.1 &  88.7/88.3 &  88.8/88.4 &  88.2/87.6\\
        EP ($\alpha=2$)   &  84.6/84.6 &  84.3/84.5 &  84.1/84.0 &  88.7/88.3 &  88.8/88.3 &  88.2/87.6\\
        EP ($\alpha=-2$)  &  84.6/84.6 &  84.3/84.5 &  83.9/84.0 &  89.0/88.7 &  89.0/88.5 &  88.1/87.6\\
        PR                &  84.6/84.6 &  84.0/84.4 &  84.0/84.1 &  88.8/88.4 &  88.8/88.4 &  88.3/87.7\\
        SW                &  84.7/84.6 &  84.2/84.5 &  84.0/84.2 &  88.8/88.4 &  88.9/88.4 &  88.2/87.6\\
        \midrule
        Value-based loc. (k=10\%) &&&&&& \\
        IP ($\alpha=0.5$)      &  84.5/84.5 &  84.4/84.4 &  84.0/84.0 &  88.9/88.5 &  88.8/88.4 &  88.2/87.7\\
        IP ($\alpha=1$)        &  84.4/84.6 &  84.2/84.4 &  84.0/84.1 &  88.8/88.4 &  88.8/88.3 &  88.2/87.6\\
        EP ($\alpha=2$)        &  84.5/84.6 &  84.5/84.3 &  84.0/84.0 &  88.7/88.2 &  88.8/88.2 &  88.1/87.6\\
        EP ($\alpha=-2$)       &  84.5/84.5 &  84.2/84.5 &  84.0/84.1 &  88.9/88.5 &  88.9/88.5 &  88.3/87.7\\
        PR                     &  \emph{83.8/84.0} &  \emph{83.5/84.0} &  \emph{83.2/83.5} &  \emph{87.6/87.3} &  \emph{86.6/86.2} &  \emph{87.2/86.8} \\
        \bottomrule
        \end{tabular}}
  \caption{\textbf{Performance of the edited stereotypical models on downstream tasks.} We fine-tune the edited models on the MNLI and STS-B tasks from the GLUE benchmark and show the results for the stereotypical target model at different sparsity levels using a single random seed. For MNLI, we report both matched and mismatched accuracy while for STS-B we, present Pearson and Spearman correlation. We emphasize the worst result per task and sparsity level. }
  \label{tab:app_glue_stereo}
\end{table*}

\begin{table*}[!t]
  \small
  \resizebox{\textwidth}{!}{
  \centering
    \begin{tabular}{lrrrrrrrr}
    \toprule
           & \multicolumn{4}{c}{MNLI-m/mm $\uparrow$} & \multicolumn{4}{c}{STS-B $\uparrow$} \\
           \cmidrule(lr){2-5} \cmidrule(lr){6-9}
           &     10\% &   20\%  &   30\%  &    40\%  &  10\% &   20\%  &   30\%  &   40\% \\
        \midrule
        \textbf{Base} & \multicolumn{4}{c}{\textbf{84.4/84.8}} & \multicolumn{4}{c}{\textbf{89.0/88.9}} \\
        \midrule
        Mask-based loc. &&&&&&&& \\
        IP ($\alpha=0.5$) &  84.5/84.7 &  84.2/84.4 &  84.0/84.4 &  83.9/84.0 &  88.6/88.2 &  88.8/88.3 &  88.5/87.9 &  88.1/87.6 \\
        IP ($\alpha=1$)   &  84.5/84.5 &  84.2/84.3 &  83.9/84.3 &  84.1/84.0 &  88.6/88.2 &  88.8/88.3 &  88.4/87.9 &  88.2/87.6 \\
        EP ($\alpha=2$)   &  84.8/84.7 &  84.2/84.3 &  83.9/84.2 &  83.9/84.1 &  88.7/88.4 &  88.9/88.4 &  88.4/87.9 &  88.1/87.6 \\
        EP ($\alpha=-2$)  &  84.5/84.7 &  84.4/84.5 &  83.7/84.3 &  84.2/84.0 &  88.5/88.1 &  88.7/88.2 &  88.3/87.8 &  88.1/87.5 \\
        PR                &  84.5/84.5 &  84.4/84.6 &  83.9/84.4 &  83.9/84.0 &  88.6/88.2 &  88.8/88.3 &  88.3/87.8 &  88.2/87.6 \\
        SW                &  84.5/84.5 &  84.2/84.4 &  84.1/84.4 &  83.9/83.9 &  88.6/88.3 &  88.8/88.3 &  88.5/87.9 &  88.1/87.6 \\
        \midrule
        Value-based loc. (k=10\%) &&&& \\
        IP ($\alpha=0.5$)      &  84.6/84.5 &  84.3/84.1 &  \emph{83.4}/84.4 &  83.8/83.8 &  88.6/88.2 &  88.8/88.3 &  88.6/88.0 &  88.1/87.5\\
        IP ($\alpha=1$)        &  84.6/84.6 &  84.3/84.6 &  84.2/84.4 &  83.9/83.8 &  88.7/88.3 &  88.8/88.3 &  88.5/88.0 &  88.1/87.5 \\
        EP ($\alpha=2$)        &  84.5/84.7 &  84.1/84.4 &  83.9/84.4 &  83.9/84.1 &  88.7/88.3 &  88.9/88.4 &  88.6/88.1 &  88.2/87.6 \\
        EP ($\alpha=-2$)       &  84.7/84.7 &  84.1/84.6 &  83.8/84.4 &  84.0/84.1 &  88.5/88.1 &  88.5/87.9 &  88.3/87.7 &  88.0/87.4 \\
        PR                     &  \emph{83.8/83.8} &  \emph{83.5/83.9} &  83.5/\emph{83.8} &  \emph{82.9/83.4} &  \emph{87.7/87.3} &  \emph{86.7/86.2} &  \emph{87.1/86.7} &  \emph{87.2/86.7} \\
        \bottomrule
        \end{tabular}}
  \caption{\textbf{Performance of the edited anti-stereotypical models on downstream tasks.} We fine-tune the edited models on the MNLI and STS-B tasks from the GLUE benchmark and show the results for the anti-stereotypical target model at different sparsity levels using a single random seed. For MNLI, we report both matched and mismatched accuracy while for STS-B we, present Pearson and Spearman correlation.}
  \label{tab:app_glue_anti}
\end{table*}

\subsection{Sensitivity to the Number of Weights Edited} \label{app:add_topk}
\begin{figure*}[t]
\centering
  \includegraphics[width=0.75\linewidth]{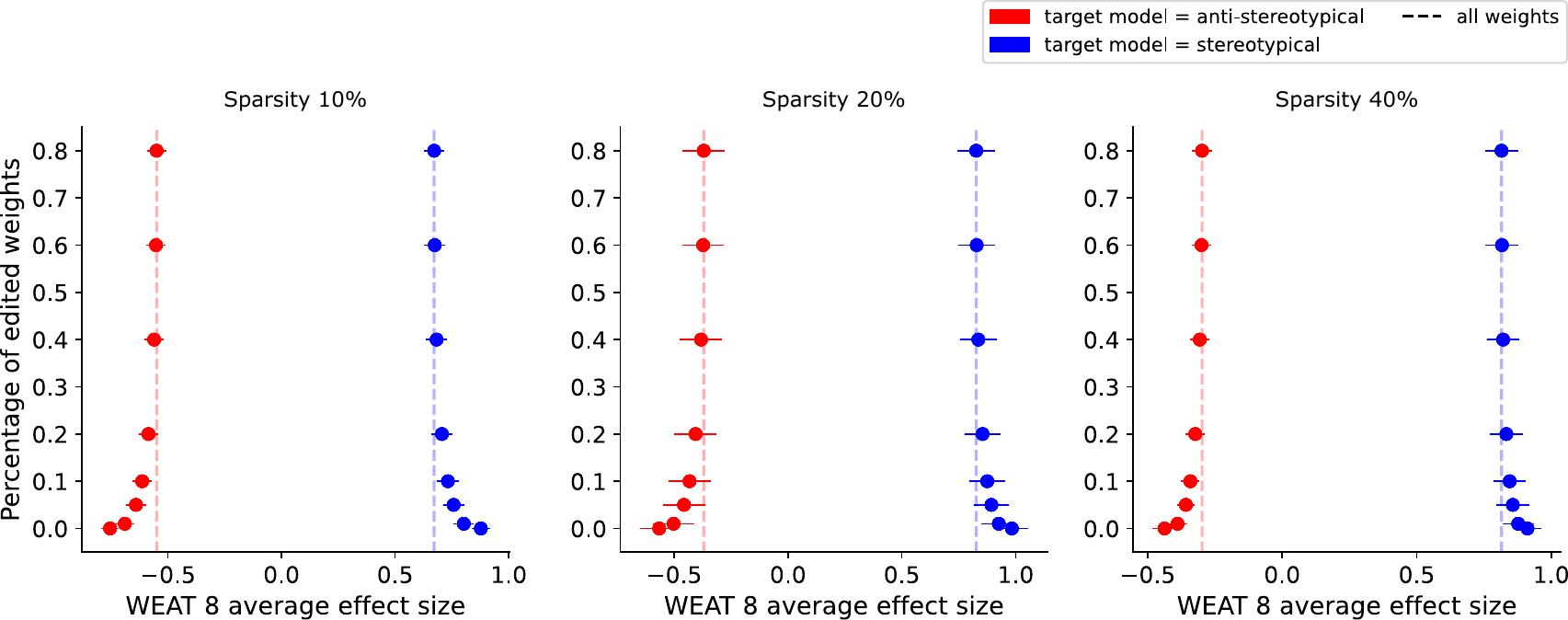} 
  \caption {\textbf{WEAT average effect size for different numbers of weights edited.} We report the mean bias at different sparsity levels across four random seeds with error bars indicating one standard deviation in each direction.} \label{fig:top_k_weat_all_sp}
\end{figure*}

\begin{figure*}[t]
\centering
  \includegraphics[width=0.75\linewidth]{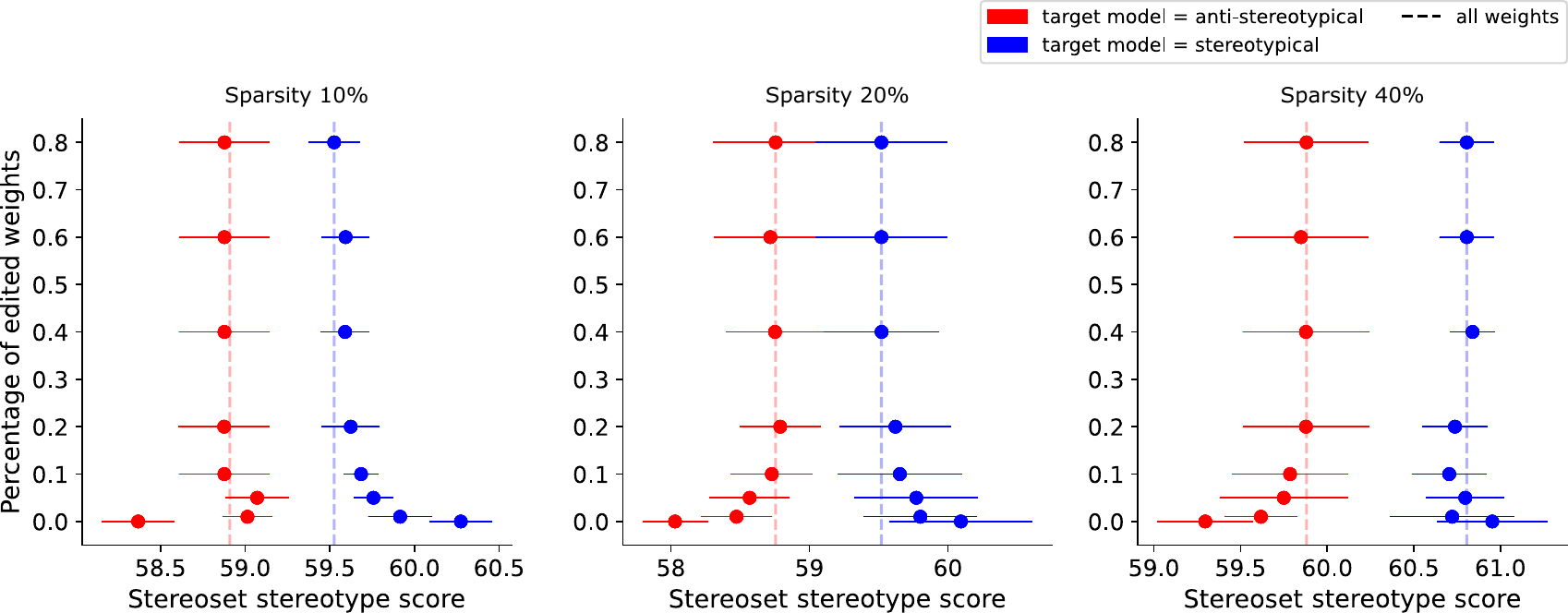} 
  \caption {\textbf{StereoSet stereotype scores for different numbers of weights edited.} We report the mean bias at different sparsity levels across four random seeds with error bars indicating one standard deviation in each direction.} \label{fig:top_k_stereoset_all_sp}
\end{figure*}

\begin{figure*}[t]
\centering
  \includegraphics[width=0.75\linewidth]{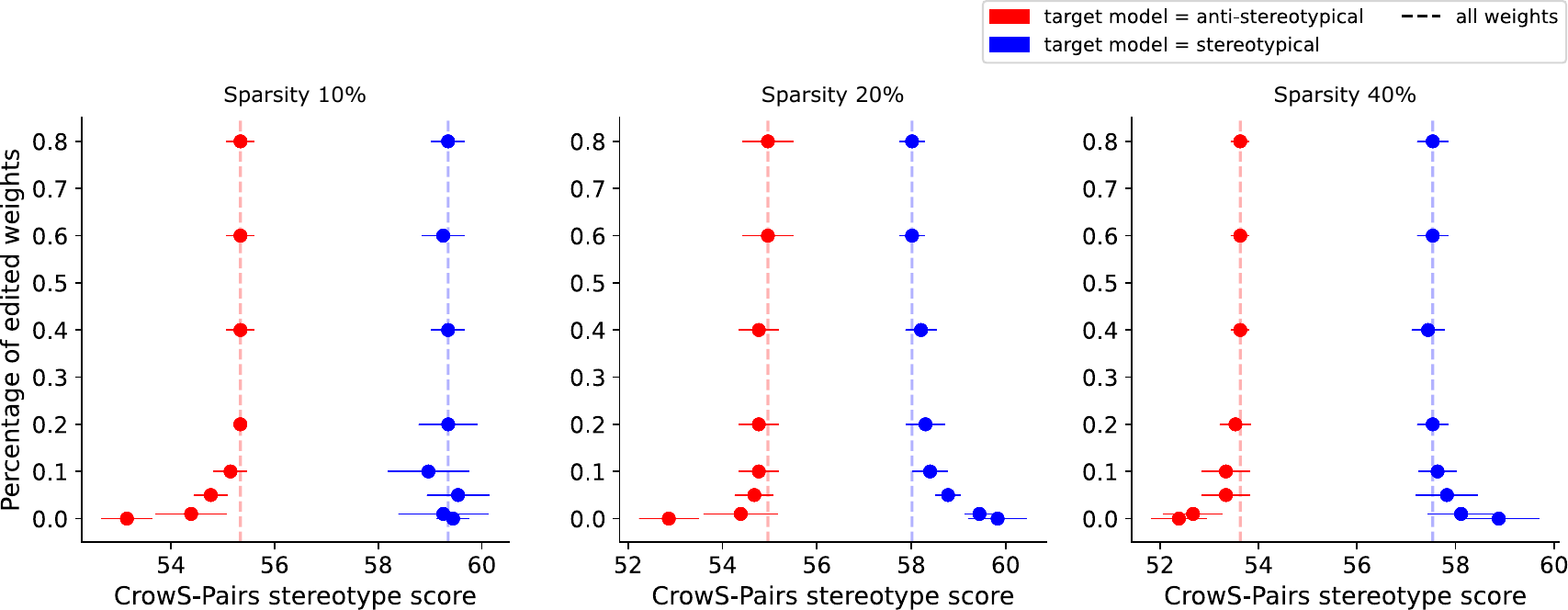} 
  \caption {\textbf{CrowS-Pairs stereotype scores for different numbers of weights edited.} We report the mean bias at different sparsity levels across four random seeds with error bars indicating one standard deviation in each direction.} \label{fig:top_k_crows_all_sp}
\end{figure*}

Figures \ref{fig:top_k_weat_all_sp}, \ref{fig:top_k_stereoset_all_sp} and \ref{fig:top_k_crows_all_sp} present the results of experiments exploring various numbers of weights selected for interpolation for additional sparsity levels, using a fixed weighting factor of ${\alpha=0.5}$.
We observe similar trends across all bias measures and sparsity levels, indicating that editing $20\% - 40\%$ of the weights has an equivalent effect on bias as editing all the weights.

\subsection{Sensitivity to the Weighting Factor} \label{app:add_alpha}

\begin{figure*}[t]
\centering
  \includegraphics[width=0.85\linewidth]{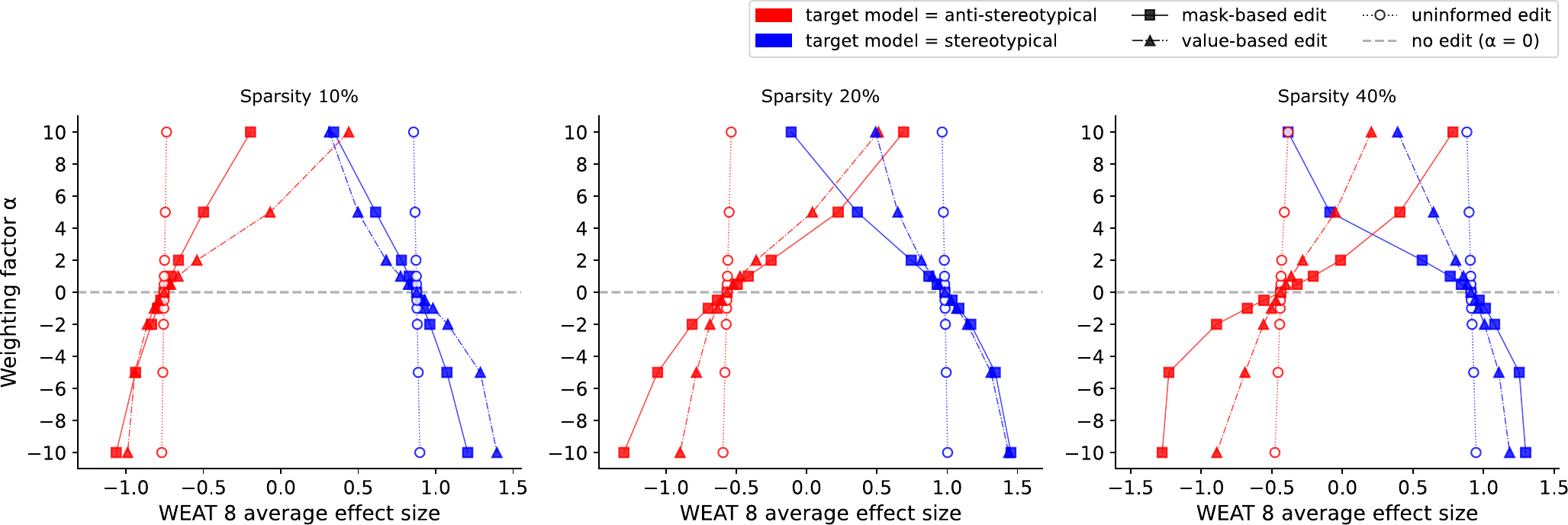} 
  \caption {\textbf{WEAT average effect size for different weighting factors.} We set the number of edited weights to the number of weights selected by masked-based localization and report the mean bias across four random seeds.} \label{fig:alpha_weat_all_sp}
\end{figure*}

\begin{figure*}[t]
\centering
  \includegraphics[width=0.85\linewidth]{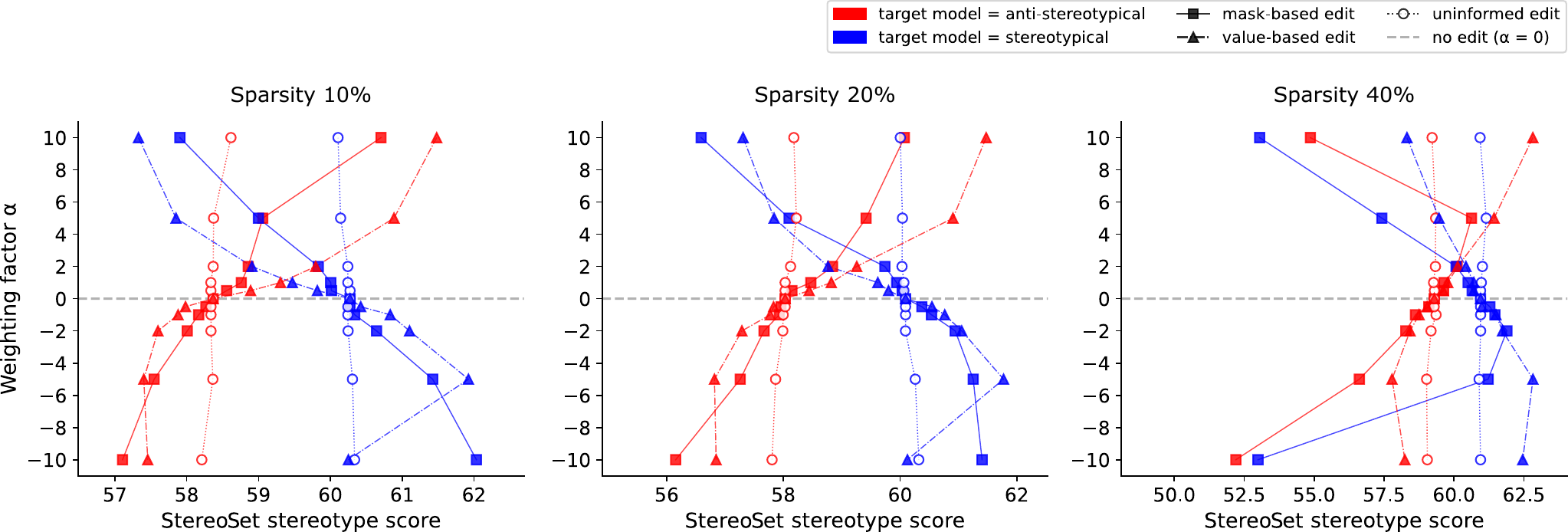} 
  \caption {\textbf{StereoSet stereotype scores for different weighting factors.} We set the number of edited weights to the number of weights selected by masked-based localization and report the mean bias across four random seeds.} \label{fig:alpha_stereoset_all_sp}
\end{figure*}

\begin{figure*}[t]
\centering
  \includegraphics[width=0.85\linewidth]{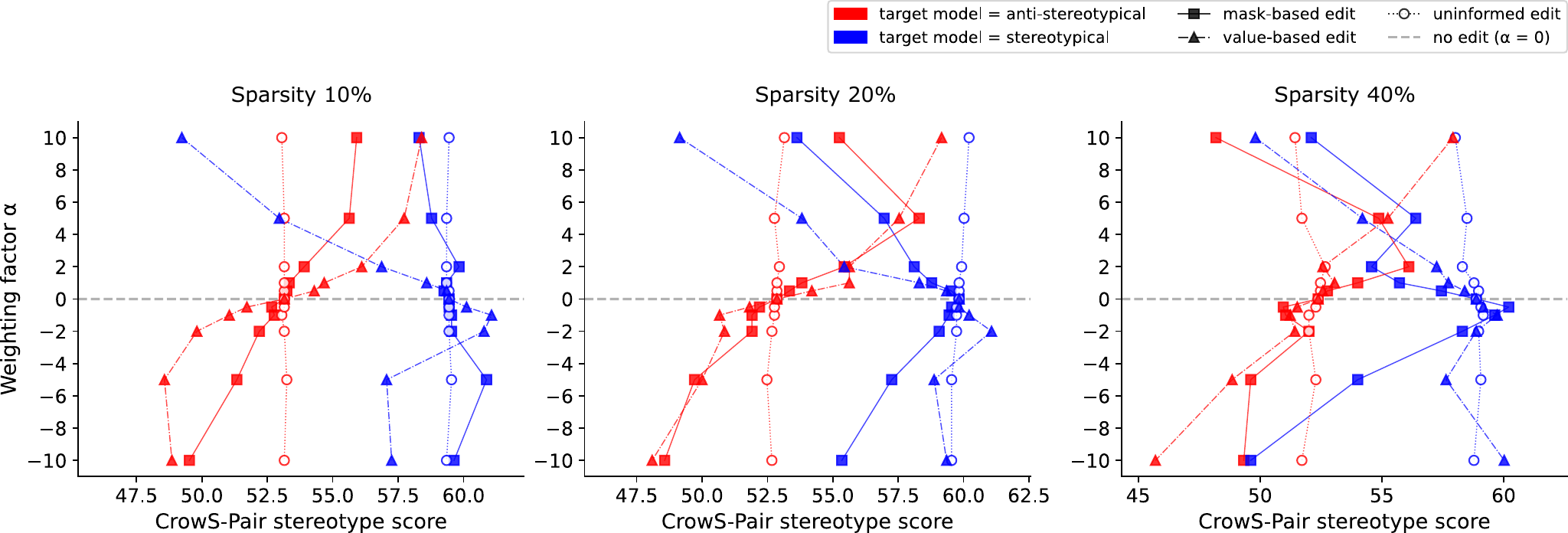} 
  \caption {\textbf{CrowS-Pairs stereotype scores for different weighting factors.} We set the number of edited weights to the number of weights selected by masked-based localization and report the mean bias across four random seeds.} \label{fig:alpha_crows_all_sp}
\end{figure*}

\begin{figure*}[t]
\centering
  \includegraphics[width=0.85\linewidth]{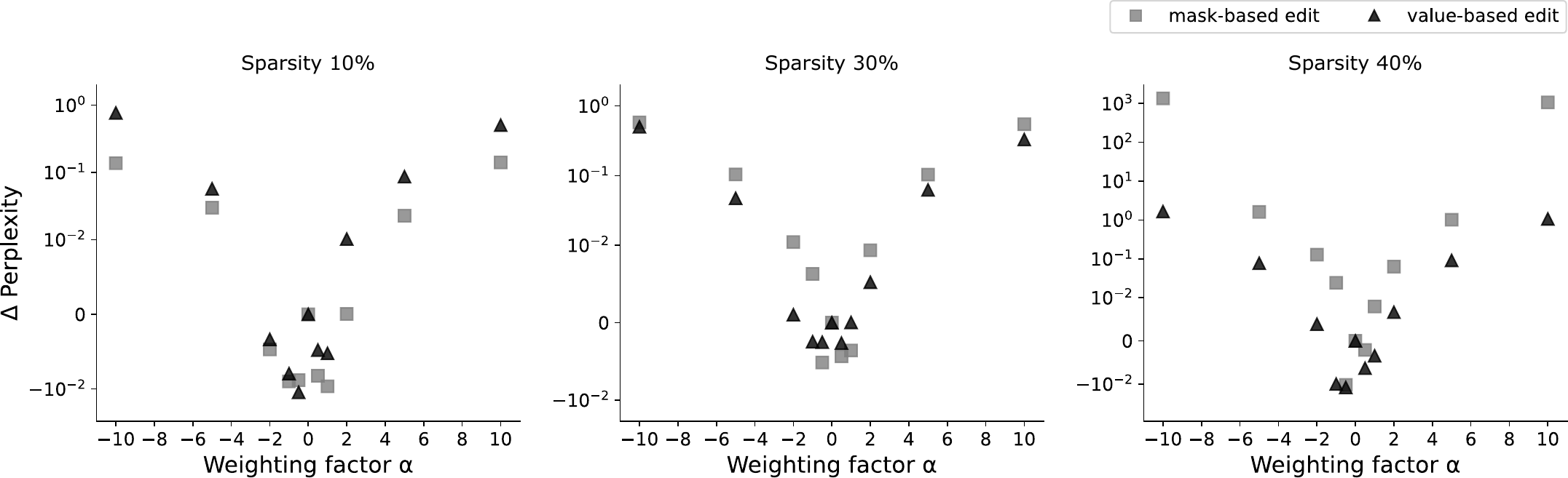} 
  \caption {\textbf{Change in perplexity for different weighting factors.} We set the number of edited weights to the number of weights selected by masked-based localization and report the mean perplexity over both target models (stereotypical and anti-stereotypical) and across four random seeds.} \label{fig:alpha_perplexity_all_sp}
\end{figure*}

\begin{figure*}[t]
\centering
  \includegraphics[width=\linewidth]{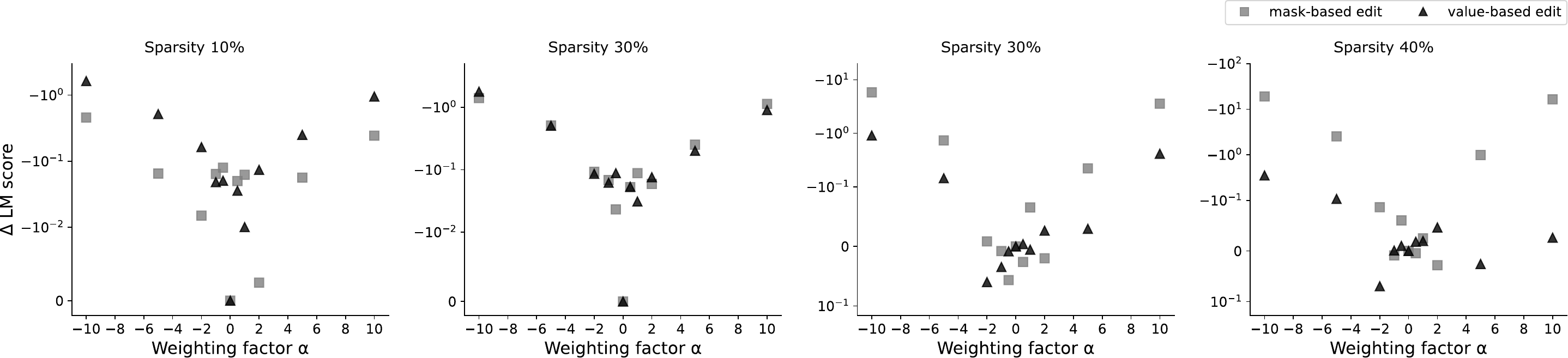} 
  \caption {\textbf{Change in LM score for different weighting factors.} We set the number of edited weights to the number of weights selected by masked-based localization and report the mean score over both target models (stereotypical and anti-stereotypical) and across four random seeds.} \label{fig:alpha_LM_all_sp}
\end{figure*}

We display results exploring different weighting factors for interpolation and extrapolation across different sparsity levels. The effect on gender bias is illustrated in figures \ref{fig:alpha_weat_all_sp}, \ref{fig:alpha_stereoset_all_sp} and \ref{fig:alpha_crows_all_sp}, while the effect on language modeling performance is shown in figures \ref{fig:alpha_perplexity_all_sp} and \ref{fig:alpha_LM_all_sp}.
We observe that across all sparsity levels and bias measures, the effect of mask-based and value-based editing on bias increases with higher absolute weighting factors (up to a certain threshold), whereas uninformed editing does not lead to any or only minor changes in bias. At the same time, perplexity and LM score indicate increasingly worse language modeling performance for higher absolute weighting factors.

 \subsection{Application to a Neutral Model} \label{app:add_neutral}

\begin{figure*}[t]
\centering
  \includegraphics[width=\linewidth]{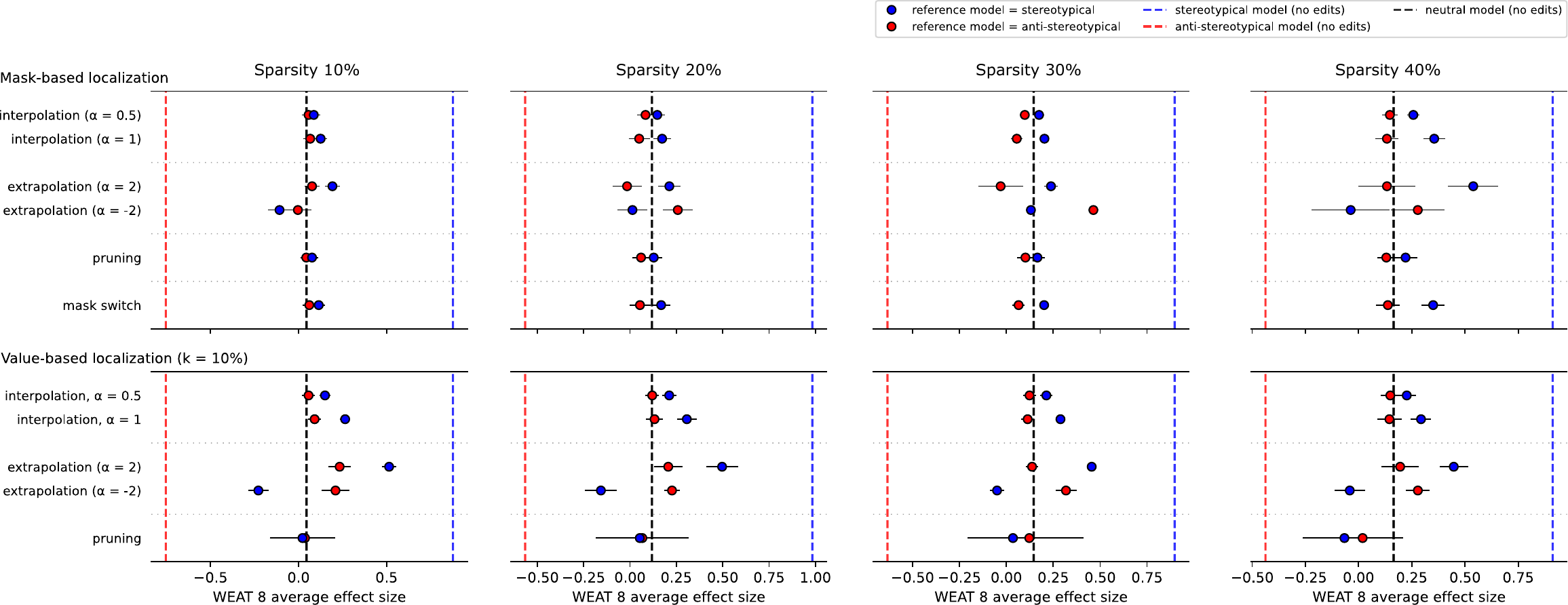} 
  \caption {\textbf{WEAT average effect size after local contrastive editing with a neutral target model.} We report the mean bias across four random seeds with error bars indicating one standard deviation in each direction.} \label{fig:neutral_weat_all_sp}
\end{figure*}

\begin{figure*}[t]
\centering
  \includegraphics[width=\linewidth]{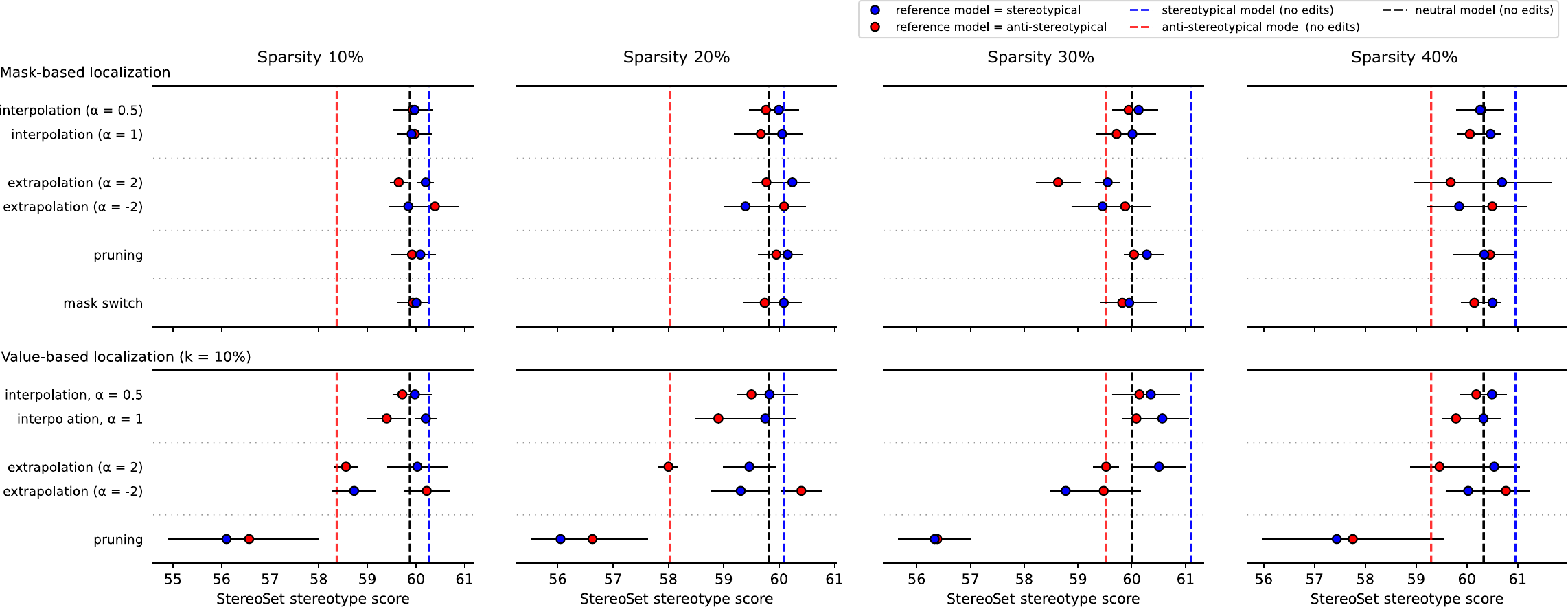} 
  \caption {\textbf{StereoSet stereotype scores after local contrastive editing with a neutral target model.} We report the mean bias across four random seeds with error bars indicating one standard deviation in each direction.} \label{fig:neutral_stereoset_all_sp}
\end{figure*}

\begin{figure*}[t]
\centering
  \includegraphics[width=\linewidth]{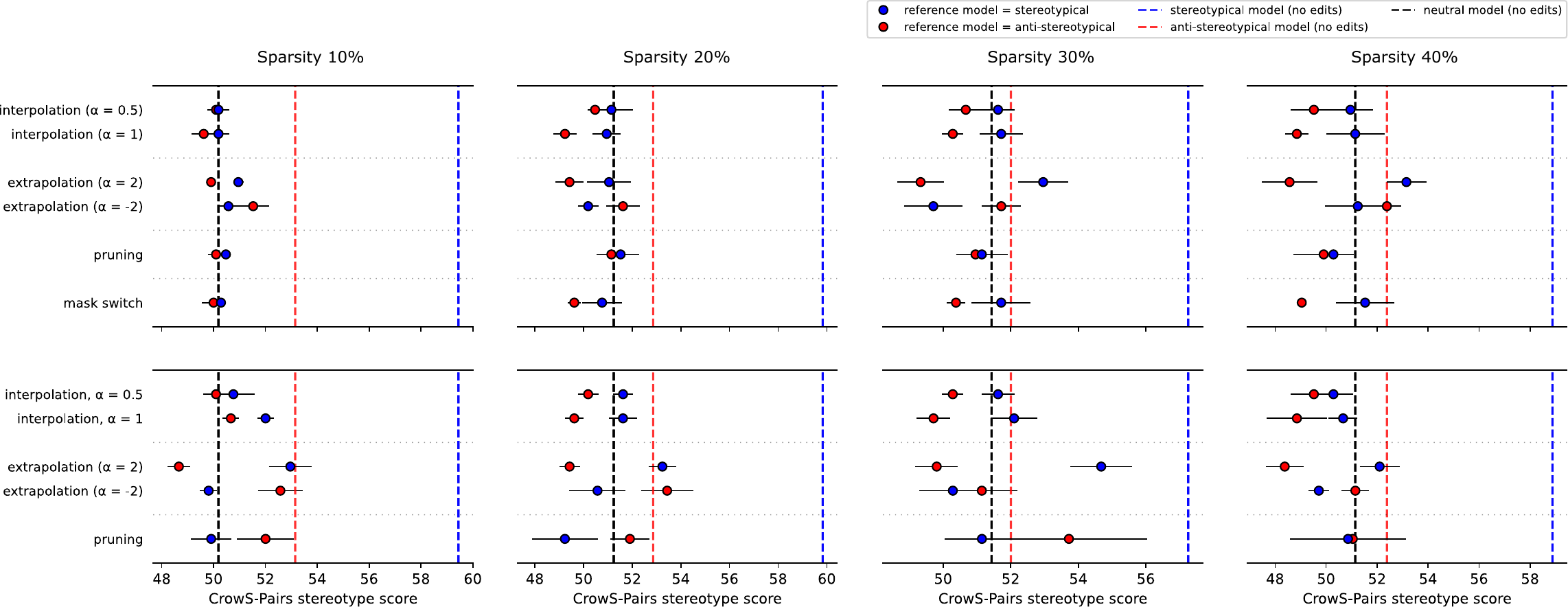} 
  \caption {\textbf{CrowS-Pairs stereotype scores after local contrastive editing with a neutral target model.} We report the mean bias across four random seeds with error bars indicating one standard deviation in each direction.} \label{fig:neutral_crows_all_sp}
\end{figure*}

We show the effect of local contrastive editing on gender bias for a neutral target model at different sparsity levels in figures \ref{fig:neutral_weat_all_sp}, \ref{fig:neutral_stereoset_all_sp} and \ref{fig:neutral_crows_all_sp}. We record the change in language modeling ability in table \ref{tab:app_perf_neutral}.

\begin{table*}[t]
  \small
  \centering
    \begin{tabular}{lrrrrrrrr}
    \toprule
           & \multicolumn{4}{c}{$\Delta$ perplexity $\downarrow$} & \multicolumn{4}{c}{$\Delta$ LM score $\uparrow$} \\
           \cmidrule(lr){2-5} \cmidrule(lr){6-9}
           &     10\% &   20\%  &   30\% &    40\%  &  10\% &      20\% &   30\%  &  40\%\\
        \midrule
        Mask-based localization &&&& \\
        IP ($\alpha=0.5$) &     +0.005          &   -0.005          &   +0.011            &   +0.013          & +0.076          & +0.238 & -0.072 & +0.062 \\
        IP ($\alpha=1$)   &     +0.012          &   +0.012          &   \textbf{+0.042}   &   \textbf{+0.055} & +0.030          & +0.164 & -0.160 & -0.073 \\
        EP ($\alpha=2$)   &     \textbf{+0.025} &   \textbf{+0.065} &   \textbf{+0.157}   &   \textbf{+0.310} & +0.109          & +0.027 & -0.250 & -0.180 \\
        EP ($\alpha=-2$)  &     \textbf{+0.027} &   \textbf{+0.076} &   \textbf{+0.221}   &   \textbf{+0.534} & \textbf{-0.419} & -0.371 & -0.122 & -0.321 \\
        PR                &     +0.006          &   +0.009          &   \textbf{+0.028}   &   \textbf{+0.040} & +0.003          & +0.059 & -0.156 & +0.0067 \\
        SW                &     \textbf{+0.011} &   \textbf{+0.015} &   \textbf{+0.041}   &   \textbf{+0.059} & +0.057          & +0.179 & -0.184 & -0.057 \\
        \midrule
        Value-based localization ($k=10\%$) &&&& \\
        IP ($\alpha=0.5$)      &   +0.006           &   -0.001          &   +0.011          &   +0.005          & -0.149          & +0.140          & -0.177          & -0.081 \\
        IP ($\alpha=1$)        &   +0.004           &   +0.005          &   \textbf{+0.017} &   +0.018          & -0.256          & +0.138          & -0.195          & \textbf{-0.177} \\
        EP ($\alpha=2$)        &   \textbf{+0.036}  &   \textbf{+0.031} &   \textbf{+0.043} &   \textbf{+0.063} & -0.393          & -0.085          & -0.141          & \textbf{-0.393} \\
        EP ($\alpha=-2$)       &   \textbf{+0.065}  &   \textbf{+0.050} &   \textbf{+0.069} &   \textbf{+0.070} & +0.153          & -0.001          & -0.053          & +0.090 \\
        PR                     &  \textbf{+4.826}   &  \textbf{+5.873}  &  \textbf{+5.194}  &  \textbf{+6.215} & \textbf{-2.682} & \textbf{-1.957} & \textbf{-1.100}  & -0.995 \\
        \bottomrule
        \end{tabular}
  \caption{\textbf{Change in language modeling ability after local contrastive editing with a neutral target model.} We show the mean change in perplexity and LM scores after local contrastive editing across both reference models (stereotypical and anti-stereotypical) and four random seeds at different sparsity levels. We print significant differences bold.}
  \label{tab:app_perf_neutral}
\end{table*}

\end{document}